\pdfoutput=1
\documentclass{article}

\usepackage[final]{neurips_2024}

\usepackage{times}
\usepackage{epsfig}
\usepackage{graphicx}
\usepackage{float}
\usepackage{wrapfig}
\usepackage{amsmath,amssymb,amsthm}
\usepackage{algorithm,algorithmicx,algpseudocode}
\usepackage{bm,xspace}
\usepackage{comment}
\usepackage{verbatim}
\usepackage{multirow}
\usepackage{balance}
\usepackage{url}
\usepackage{booktabs}
\usepackage{etoolbox,siunitx}
\usepackage{calc}
\usepackage{pifont,hologo}
\usepackage{color}
\usepackage{ulem}

\newcolumntype{P}[1]{>{\centering\arraybackslash}p{#1}}
\newcolumntype{M}[1]{>{\centering\arraybackslash}m{#1}}

\usepackage[super]{nth}
\usepackage{nicefrac}
\robustify\bfseries

\DeclareMathSymbol{@}{\mathord}{letters}{"3B}

\def\latex/{\LaTeX}
\def\bibtex/{\hologo{BibTeX}}

\makeatletter
\DeclareRobustCommand\onedot{\futurelet\@let@token\@onedot}
\def\@onedot{\ifx\@let@token.\else.\null\fi\xspace}

\def\etal{et al\onedot}

\newcommand*{\Rom}[1]{\expandafter\@slowromancap\romannumeral #1@}
\newcommand*{\rom}[1]{\expandafter\romannumeral #1}

\def\eqref#1{equation~\ref{#1}}

\def\1{\bm{1}}

\DeclareMathAlphabet{\mathsfit}{\encodingdefault}{\sfdefault}{m}{sl}
\SetMathAlphabet{\mathsfit}{bold}{\encodingdefault}{\sfdefault}{bx}{n}

\usepackage[table]{xcolor}
\definecolor{blue}{HTML}{0055cc}
\definecolor{red}{HTML}{cc1100}
\definecolor{orange}{HTML}{cc7700}
\definecolor{green}{HTML}{339955}
\definecolor{Highlight}{rgb}{0.12,0.49,0.85}
\usepackage{float}
\usepackage[utf8]{inputenc} 
\usepackage[T1]{fontenc}    
\usepackage[pagebackref,breaklinks,colorlinks,linkcolor=purple,citecolor=Highlight]{hyperref}       
\usepackage{url}            
\usepackage{booktabs}       
\usepackage{amsfonts}       
\usepackage{nicefrac}       
\usepackage{microtype}      
\usepackage{xcolor}         
\usepackage{amsmath}        
\usepackage{amssymb}        
\usepackage{lipsum}         
\usepackage{bbding}         
\usepackage{array}          
\usepackage{subcaption}     
\usepackage{pdfpages}
\usepackage{natbib}
\usepackage{ulem}

\usepackage{tocbibind}
\usepackage{titleps}
\usepackage{titletoc}
\usepackage{calc}
\usepackage{etoolbox}

\usepackage{makecell}

\usepackage{caption}
\setcitestyle{square,numbers,comma,sort}
\usepackage[capitalize]{cleveref}
\crefname{section}{Sec.}{Secs.}
\Crefname{section}{Section}{Sections}
\Crefname{table}{Table}{Tables}
\crefname{table}{Tab.}{Tabs.}

\newlength\savewidth

\renewcommand{\paragraph}[1]{\noindent\textbf{#1}}

\newcolumntype{x}[1]{>{\centering\arraybackslash}p{#1pt}}
\newcolumntype{y}[1]{>{\raggedright\arraybackslash}p{#1pt}}
\newcolumntype{z}[1]{>{\raggedleft\arraybackslash}p{#1pt}}

\newcommand{\app}{\raise.17ex\hbox{$\scriptstyle\sim$}}

\definecolor{deemph}{gray}{0.6}

\definecolor{baselinecolor}{gray}{.9}

\definecolor{my_red}{HTML}{FE4444}
\definecolor{Gray}{gray}{0.95}

\newcommand{\best}[1]{{{\textcolor{my_red}{#1}}}}
\newcommand{\second}[1]{{\textcolor[HTML]{0000FF}{{#1}}}}

\let\originalleft\left
\let\originalright\right
\renewcommand{\left}{\mathopen{}\mathclose\bgroup\originalleft}
\renewcommand{\right}{\aftergroup\egroup\originalright}

\title{Parameter Efficient Adaptation for Image Restoration \\with Heterogeneous Mixture-of-Experts}

\author{%
    Hang Guo$^1$ \quad 
    Tao Dai\footnotemark[1] $\ ^2$ \quad 
    Yuanchao Bai$^3$ \quad 
    Bin Chen$^3$ \\ 
    \textbf{Xudong Ren$^1$ \quad 
    Zexuan Zhu$^2$ \quad 
    Shu-tao Xia$^{1,4}$} 
  \vspace{0.1cm} \\
  $^1$Tsinghua University \quad
  $^2$Shenzhen University \\
  $^3$Harbin Institute of Technology \quad 
  $^4$Pengcheng Larboratory \quad
  \vspace{0.1cm} \\
  \url{https://github.com/csguoh/AdaptIR}
}

\begin{document}
\renewcommand{\thefootnote}{\fnsymbol{footnote}}
\footnotetext[1]{Corresponding author: Tao Dai}
\maketitle

\vspace{-6mm}
\begin{abstract}
\vspace{-2mm}

Designing single-task image restoration models for specific degradation has seen great success in recent years. To achieve generalized image restoration, all-in-one methods have recently been proposed and shown potential for multiple restoration tasks using one single model. Despite the promising results, the existing all-in-one paradigm still suffers from high computational costs as well as limited generalization on unseen degradations. In this work, we introduce an alternative solution to improve the generalization of image restoration models. Drawing inspiration from recent advancements in Parameter Efficient Transfer Learning (PETL), we aim to tune only a small number of parameters to adapt pre-trained restoration models to various tasks. However, current PETL methods fail to generalize across varied restoration tasks due to their homogeneous representation nature. To this end, we propose AdaptIR, a Mixture-of-Experts (MoE) with orthogonal multi-branch design to capture local spatial, global spatial, and channel representation bases, followed by adaptive base combination to obtain heterogeneous representation for different degradations. Extensive experiments demonstrate that our AdaptIR achieves stable performance on single-degradation tasks, and excels in hybrid-degradation tasks, with fine-tuning only 0.6\% parameters for 8 hours.

\end{abstract}
\vspace{-2mm}

\section{Introduction}
Image restoration, aiming to restore high-quality images from their degraded counterparts, is a fundamental computer vision problem and has been studied for many years. Due to its ill-posed nature, early research efforts~\cite{dong2014learning,zhang2017beyond,ren2019progressive} typically focus on developing single-task models, with each model handling only one specific degradation. Consequently, these methods often exhibit limited generalization across different image restoration tasks.

To improve generalization ability, all-in-one image restoration methods~\cite{li2022all,potlapalli2023promptir,park2023all} have recently been proposed and have attracted great research interest. By training one model with multiple degradation data, these methods enable the single model to handle various degradations. Despite the promising results, the existing all-in-one paradigm still faces several challenges. Firstly, the all-in-one model can only restore degradations encountered during training; once training is complete, the model cannot handle new degradations. Secondly, since the knowledge of restoring multiple degradations is learned by a single model, it incurs a significant cost to train and store these all-in-one models.

In this work, we propose an alternative solution to improve the generalization ability of restoration models in handling multiple degradations. Drawing inspiration from Parameter Efficient Transfer Learning (PETL)~\cite{chen2022adaptformer,lester2021power,jia2022visual,jie2023fact}, we aim to insert a small number of trainable modules into frozen pre-trained restoration backbones. By training only these newly added modules on downstream tasks, the pre-trained restoration backbone can be adapted to unseen restoration tasks. Since only a small number of parameters need to be trained, the training cost is very small and the training process can converge quickly when new tasks are added. When the training is completed, only the newly added parameters need to be stored, thus greatly reducing the storage cost.

\begin{figure}[!tb]
\centering
\includegraphics[width=\columnwidth]{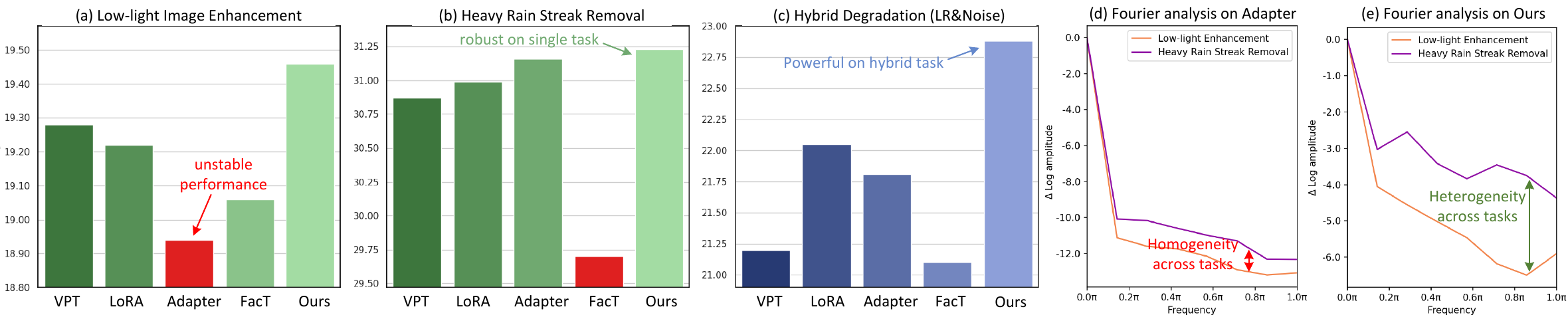}
\vspace{-6mm}
\caption{
(a)\&(b) We find directly applying the current PETL methods to image restoration leads to unstable performance on single degradation. (c) The current PETL method suffers sub-optimal results on hybrid degradation which requires heterogeneous representation. (d)\&(e) We use Fourier analysis to visualize Adapter and our AdaptIR and find that Adapter exhibits homogeneous frequency representations even when faced with different degradations, while our AdaptIR can adaptively learn degradation-specific heterogeneous representations. We provide more evidence in~\cref{sec:more-evidence}.
}
\vspace{-4mm}
\label{fig:motivation}
\end{figure}

Despite the potential of applying PETL techniques to image restoration, our experiments reveal that existing PETL methods can work normally on specific degradation, but fail to generalize across multiple degradations, exhibiting unstable performance when adapted to different restoration tasks. As shown in \cref{fig:motivation}(a), the most widely used PETL method, Adapter~\cite{lester2021power}, performs well on the draining task. However, when applying Adapter to the low-light image enhancement task, the adapted model shows significant performance degradation. This phenomenon also occurs with other methods, such as the recent state-of-the-art PETL method FacT~\cite{jie2023fact}(\cref{fig:motivation}(b)). This confusing phenomenon motivates us to discover possible reasons.

To this end, we design preliminary experiments, in which we fine-tune the pre-trained restoration model~\cite{chen2021pre} using existing PETL schemes, and then use Fourier analysis~\cite{park2021vision} to observe the frequency characteristics of features from these methods. It is observed that the features from current PETL methods exhibit homogeneous representation across different restoration tasks (see \cref{fig:motivation}(d)). As demonstrated in previous work~\cite{park2023all}, different restoration tasks prefer certain representations for optimal results, we thus hypothesize that the performance drop occurs 
when the representation needed to address one specific degradation does not match the homogeneous representation of existing PETL methods. To verify this hypothesis, we further test current PETL methods using the hybrid degradation task (\cref{fig:motivation}(c)), which requires heterogeneous representations to handle diverse degradations, and we find all existing approaches suffer severe performance drops. Based on the above experiments, we argue that the homogeneous representation of existing PETL methods hinders stable performance on single degradation tasks and advanced performance on hybrid degradation tasks.

In order to learn heterogeneous representations across tasks, one possible solution is the multi-branch structures, where each branch is designed to learn orthogonal representation bases, and then adaptively combine these bases for specific degradation. Following this idea, we propose AdaptIR, a heterogeneous Mixture-of-Experts (MoE) to adapt pre-trained restoration models with heterogeneous representations across tasks. Our AdaptIR adopts orthogonal multi-branch design to learn local spatial, global spatial, and channel representation bases. Specifically, The Local Interaction Module (LIM) employs depth-separable convolution with kernel weight decomposition to exploit local spatial representation. We then employ the Frequency Affine Module (FAM), which performs frequency affine transformation to introduce global spatial modeling ability. Additionally, the Channel Gating Module (CGM) is adopted to capture channel interactions. Finally, we utilize the Adaptive Feature Ensemble to dynamically fuse these three representation bases for specific degradation.
Thanks to the heterogeneous representation modeling, our AdaptIR achieves stable performance on single-degradation tasks and advanced performance on hybrid-degradation tasks.

The contributions of our work are as follows:
\textbf{(i)} We propose a novel PETL paradigm to improve the generalization for image restoration models, and further investigate the specific challenges when applying existing PETL methods to low-level restoration.
\textbf{(ii)} We introduce AdaptIR, a custom PETL method that employs a heterogeneous MoE for orthogonal representation modeling. To the best of our knowledge, this is the first work to explore parameter-efficient adaptation for image restoration.
\textbf{(iii)} Experiments on various downstream tasks demonstrate that AdaptIR achieves robust performance on single degradation tasks and advanced results on hybrid degradation tasks.

\section{Related Work}
\label{sec:related work}

\subsection{Generalized Image Restoration}
Image restoration has attracted a lot of research interest in recent years. Due to the challenging ill-posed nature, some early research paradigms typically study each sub-task in image restoration independently and have recently achieved favorable progress in their respective fields~\cite{zhang2017learning,dai2019second,yang2019joint,zhang2017beyond}. However, designing such a single-task model is cumbersome, and does not consider the similarities among different tasks. Recently, all-in-one image restoration~\cite{li2022all,potlapalli2023promptir,park2023all} has offered a way to improve the generalization of image restoration models. By training one single model on multiple degradations, it allows the model to have the ability to handle multiple degradations. For example, AirNet~\cite{li2022all} proposes a two-stage training scheme to first learn the degradation representation, which is then used in the following restoration stage. PromptIR~\cite{potlapalli2023promptir} utilizes prompt learning to obtain degradation-specific prompts to train the model in an end-to-end manner. Despite the progress, the current all-in-one restoration paradigm can only deal with degradation seen during training and it is inevitable to re-train the model when needing to add new degradations. In addition, incorporating the knowledge of handling multiple degradations into one model has to increase the model size, which causes large training and storage costs.

\subsection{Parameter-Efficient Transfer Learning}
Parameter efficient transfer learning, which initially came from with NLP~\cite{houlsby2019parameter,zaken2021bitfit,liu2023pre,lester2021power,li2021prefix, hu2021lora,karimi2021compacter,he2021towards}, aims to catch up with full fine-tuning by training a small number of parameters. Recently, this technique has emerged in the field of computer vision with promising results~\cite{jia2022visual,chen2022adaptformer,lian2022scaling,jie2023fact,chavan2023one,zhang2022NOAH,chen2022vision,liu2022polyhistor,jie2022convolutional,jie2023revisiting}. For example, VPT~\cite{jia2022visual} adds learnable tokens, also called prompts, to the input sequence of one frozen transformer layer. Adapter~\cite{houlsby2019parameter} employs a bottleneck structure to adapt the pre-trained model. Some attempts also introduce parameterized hypercomplex multiplication layers~\cite{karimi2021compacter} and re-parameterisation~\cite{luo2023towards} to adapter-based methods. Moreover, LoRA~\cite{hu2021lora} utilizes the low-rank nature of the incremental weight in attention and performs matrix decomposition for parameter efficiency. He \etal ~\cite{he2021towards} go further to identify all the above three approaches from a unified perspective. In addition, NOAH~\cite{zhang2022NOAH} and GLoRA~\cite{chavan2023one} introduce Neural Architecture Search (NAS) to combine different methods. SSF~\cite{lian2022scaling} performs a learnable affine transformation on features of the pre-trained model. 
FacT~\cite{jie2023fact} tensorizes ViT and decomposes the increments into lightweight factors. Although applying PETL methods to pre-trained image restoration models to improve the generalization seems promising, we find that current PETL methods suffer from homogeneous representations when facing different degradations, hindering stable performance across tasks.

\section{Method}

\subsection{Preliminary}
\begin{figure*}[!tb]
\centering
\includegraphics[width=\textwidth]{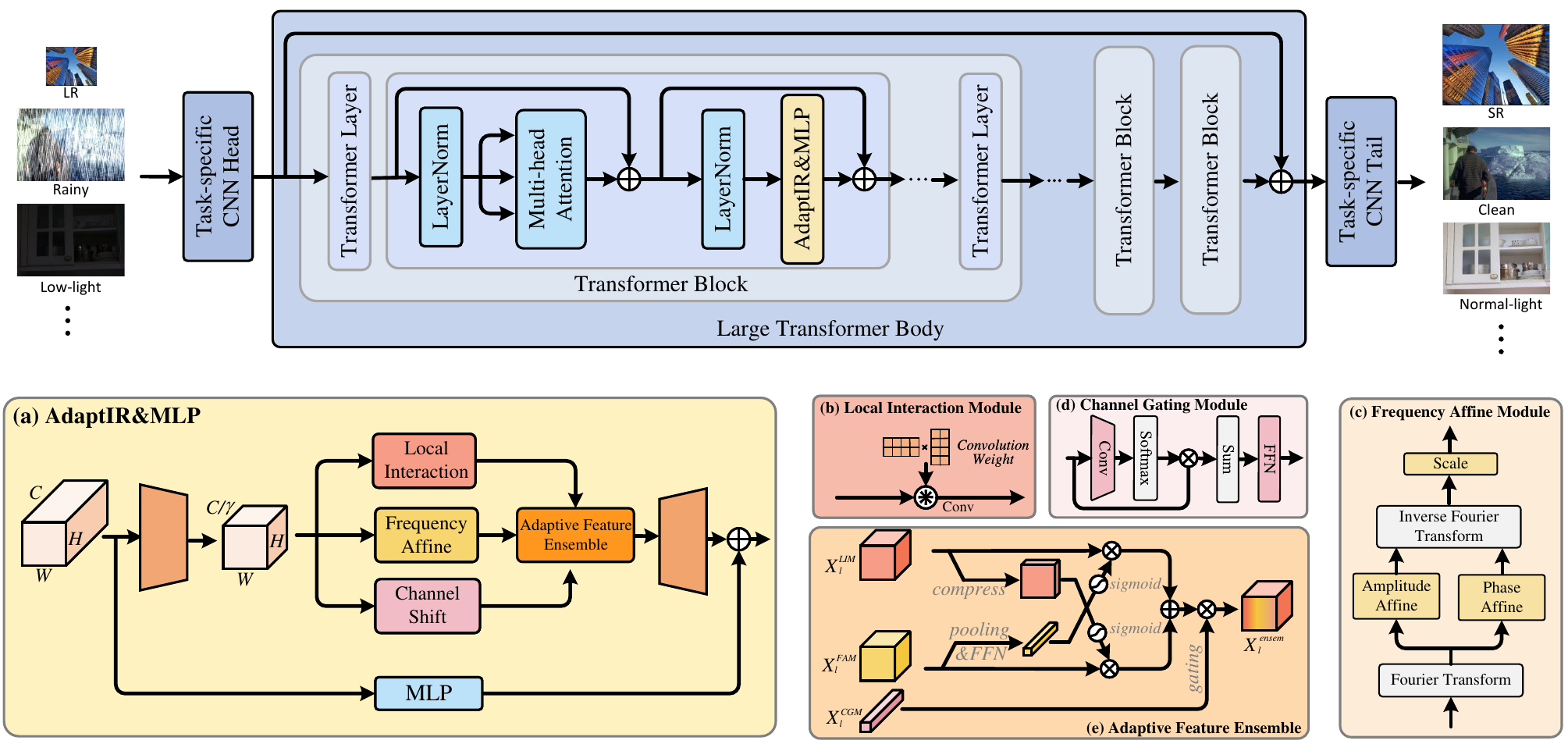}
\caption{An illustration of the proposed AdaptIR. Our AdaptIR is placed parallel to the frozen MLP in one transformer layer and thus can be seamlessly inserted into various transformer-based pre-trained restoration models.}
\label{fig:pipeline}
\vspace{-3mm}
\end{figure*}

In this work, we aim to adapt pre-trained restoration models to multiple downstream tasks by fine-tuning a small number of parameters. Following existing PETL works, we mainly focus on transformer-based restoration models since transformer has been shown to be suitable for pre-training~\cite{achiam2023gpt4} and there is no CNN-based pre-trained model available.
As shown in \cref{fig:pipeline}, a typical pre-trained restoration model~\cite{chen2021pre,li2021efficient} usually contains one large transformer body as well as task-specific heads and tails. Given the pre-assigned task type, the low-quality image $I_{LQ}$ will first go through the corresponding head to get the shallow feature $X_{head}$. After that, $X_{head}$ is flattened into a 1D sequence on the spatial dimension and is input to the transformer body which contains several stacked transformer blocks with each block containing multiple transformer layers~\cite{vaswani2017attention}. Finally, a skip connection is adopted followed by the task-specific tail to reconstruct the high-quality image $I_{HQ}$. During the pre-training stage, gradients from multiple tasks are used to update the shared body as well as the corresponding task-specific head and tail. After pre-training the restoration model, previous common practice fine-tunes all parameters of the pre-trained model for specific downstream tasks, which burdens training and storage due to the per-task model weights.

\subsection{Heterogeneous Representation Learning}

To obtain stable performance across multiple restoration tasks, it is crucial to allow the learning of heterogeneous representation for different degradations. To this end, we formalize AdaptIR as a multi-branch MoE structure, where each branch learns representations orthogonal to each other to form the representation bases, and then these bases are adaptively combined to achieve degradation-specific representation. Formally, as shown in~\cref{fig:pipeline}(a), since the transformer body flattens the $l$-th layer feature into a 1D token sequence, we first restore the 2D image structure to obtain $X_l \in \mathbb{R}^{C \times H \times W}$. After that, we apply the $1 \times 1$ convolution with channel reduction rate $\gamma$ to transfer $X_l$ to low-dimension space for parameter efficiency and obtain the intrinsic feature ${X}_l^{intrin} \in \mathbb{R}^{\frac{C}{\gamma} \times H \times W}$. Then three parallel branches are orthogonally designed to learn local spatial, global spatial, and channel bases. Next, bases from these three branches are adaptively ensembled to obtain representation $X_l^{adapt}$ for specific degradation. Finally, $X_l^{adapt}$ is added to the output of the frozen MLP to adapt the pre-trained restoration models. Details of the three branches are given below.

\noindent \textbf{Local Interaction Module.} 
We first introduce the Local Interaction Module (LIM) to model the local spatial representation. As shown in~\cref{fig:pipeline}(b), the proposed LIM is implemented by the depth-wise convolution with weight factorization for parameter efficiency. Specifically, given the convolution weight $W \in \mathbb{R}^{C_{in} \times \frac{C_{out}}{group} \times K \times K}$, where $C_{in}$, $C_{out}$ are input and output channel, $K$ is the kernel size and $group$ is the number of convolution groups, we first reshape $W$ into a 2D weight matrix $W' \in \mathbb{R}^{C_{in} \times \frac{C_{out}}{group} K^2}$, and then decompose $W'$ into multiplication of two low-rank weight matrices:
\begin{equation}
    W' = U  V^{\top},
\end{equation}
\noindent
where $U \in \mathbb{R}^{C_{in} \times r}$, $V \in \mathbb{R}^{\frac{C_{out}}{group} K^2 \times r}$ and $r$ is the rank to trade-off performance and efficiency. Then we reshape $W'$ to the original kernel size and use it to convolve $X_l^{intrin}$ to get $X_l^{LIM}$: 
\begin{equation}
    X_l^{LIM} = {\rm Reshape}(W')\circledast X_{l}^{intrin},
\end{equation}
\noindent
where $\circledast$ denotes convolution operator, and $\rm{Reshape}(\cdot)$ transforms 2D matrices into 4D convolution kernel weights. 

\vspace{1mm}
\noindent \textbf{Frequency Affine Module.}
We then consider modeling global spatial to achieve orthogonal spatial modeling to LIM. A possible solution is to introduce the attention mechanism~\cite{vaswani2017attention} which has a global receptive field. However, the attention comes at the cost of high complexity, which goes against the principle of parameter efficiency. In this work, we resort to the frequency domain for a solution.
Specifically, we apply the Fast Fourier Transform (FFT) on $ X_l^{intrin}$ to obtain the corresponding frequency feature map $X_l^\mathcal{F} \in \mathbb{C}^{ \frac{C}{\gamma} \times H \times (\lfloor \frac{W}{2} \rfloor +1)}$:
\begin{equation}
    X_l^\mathcal{F}(u,v) = \frac{1}{HW} \sum_{h=0}^{H-1} \sum_{w=0}^{W-1} X^{intrin}_l(h,w)e^{-2 \pi i (\frac{uh}{H}+\frac{vw}{W})},
\label{eq:fft}
\end{equation}
\vspace{1mm}
\noindent
As can be seen from Eq.\ref{eq:fft}, a good property of FFT is that each position of the frequency feature map is the weighted sum of all features in the spatial domain. Therefore, performing pixel-wise projection on $X_l^\mathcal{F}$ is equivalent to performing a global operator in the spatial domain.

Motivated by this observation, we propose the Frequency Affine Module (FAM) to take advantage of the inherent global representation in $X_l^\mathcal{F}$ (see \cref{fig:pipeline}(c)). Concretely, we perform the affine transformation on amplitude map $Mag_l$ and phase map $Pha_l$ respectively with depth-separable $ 1 \times 1 $ convolution. To ensure numerical stability during the early training stages, we initialize the transformation layers as all-one weights and zero bias. Subsequently, the inverse Fast Fourier Transform (iFFT) is applied to convert the affined feature back to the spatial domain. Finally, another depth-separable $ 1 \times 1 $ convolution is used as a scale layer for subsequent feature ensemble. In short, the whole process can be formalized as:
\begin{equation}
\begin{aligned}
&  [Mag_l, Pha_l] = {\rm FFT}(X_l^{intrin}),\\
& X_l^{FAM} = {\rm Conv}({\rm iFFT}({\rm to\_complex}(\phi_1 (Mag_l),\phi_2 (Pha_l)))),
\end{aligned}
\end{equation}
\noindent
where $\phi_1(\cdot)$ and $\phi_2(\cdot)$ are the frequency projection function and $\rm {to\_complex( \cdot, \cdot)}$ converts the magnitude and phase to complex numbers.

\noindent \textbf{Channel Gating Module.}
The above LIM and FAM both adopt the depth-separable strategy for parameter efficiency. To allow for another orthogonal representation, we further develop the Channel Gating Module (CGM) for salient channel selection. As shown in \cref{fig:pipeline}(d), we first obtain the spatial weight mask $\mathcal{M}_l \in \mathbb{R}^{1 \times H \times W}$ by employing $1 \times 1$ convolution which compresses the channel dimension of $X_l^{intrin}$ to 1, followed by the $\mathrm{Softmax}$ on the spatial dimension:
\begin{equation}
    \mathcal{M}_l = {\rm Softmax}({\rm Conv}(X_l^{intrin})).
\end{equation}
\noindent
We then apply $\mathcal{M}_l$ on each channel of $X_l^{intrin}$ to perform spatially weighted summation to obtain the channel vector which will go through a Feed Forward Network (FFN) to generate the channel gating factor $X_l^{CGM} \in \mathbb{R}^{\frac{C}{\gamma}\times 1 \times 1}$:
\begin{equation}
X_l^{CGM} = {\rm FFN}(\sum_{h,w}\mathcal{M}_l\otimes X_l^{intrin}),
\end{equation}
\noindent
where $\otimes$ denotes the Hadamard product.

\vspace{1mm}
\noindent \textbf{Adaptive Feature Ensemble.}
The orthogonal modeling from local spatial, global spatial, and channel can serve as favorable representation bases, and we further introduce the Adaptive Feature Ensemble to learn their combinations to obtain degradation-specific representations. As shown in \cref{fig:pipeline}(e), we use convolution to compress the channel of $X_l^{LIM}$ to 1 to extract its spatial details, while applying global average pooling and FFN on $X_l^{FAM}$ to preserve the global information. Next, the sigmoid activation is employed to generate dynamic weights for multiplication and produces the adaptive spatial features $X_l^{spatial}$. After that, we use $X_l^{CGM}$ to perform channel selection on $X_l^{spatial}$ to obtain the degradation-specific ensemble feature $X_l^{ensem} \in \mathbb{R}^{\frac{C}{\gamma} \times H \times W}$.

At last, a $1\times1$ convolution is employed to up-dimension the $X_l^{ensem}$ to generate $X_l^{adapt}\in\mathbb{R}^{C \times H \times W}$. For stability, we use zero to initialize the convolution weights. Then, the $X_l^{adapt}$ is added to the output of frozen MLP as residual to adapt pre-trained models to downstream tasks.

\subsection{Parameter Efficient Training}

During training, we freeze all parameters in the pre-trained model, including task-specific heads and tails as well as the transformer body except for the proposed AdaptIR. A simple $L_1$ loss is employed to provide pixel-level supervision:
\begin{equation}
    \mathcal{L}_{pix} = ||I_{HQ}-I_{LQ}||_1,
\end{equation}
\noindent
where $||\cdot||_1$ denotes $L_1$ norm.

\section{Experiment}
We first employ single-degradation restoration tasks to assess the performance stability of different PETL methods, including image SR, color image denoising, image deraining, and low-light enhancement. Subsequently, we introduce hybrid degradation to further evaluate the ability to learn heterogeneous representations.
In addition, we compare with recent all-in-one methods in both effectiveness and efficiency to demonstrate the advantages of applying PETL for generalized image restoration.  Finally, we conduct ablation studies to reveal the working mechanism of the proposed method as well as different design choices. Since evaluating the performance stability requires experiments on multiple single-degradation tasks, due to the page limit, the related experiments can be seen in the~\cref{sec:single_task}.

\vspace{-3mm}
\subsection{Experimental Settings}

\noindent \textbf{Datasets.} 

For image SR, we choose DIV2K~\cite{agustsson2017ntire} and Flickr2K~\cite{timofte2017ntire} as the training set, and we evaluate on Set5~\cite{bevilacqua2012low}, Set14~\cite{zeyde2012single}, BSDS100~\cite{arbelaez2010contour}, Urban100~\cite{huang2015single}, and Manga109~\cite{matsui2017sketch}. For color image denoising, training sets consist of DIV2K ~\cite{agustsson2017ntire}, Flickr2K~\cite{timofte2017ntire}, BSD400~\cite{arbelaez2010contour}, and WED~\cite{ma2016waterloo}, and we have two testing sets: CBSD68~\cite{martin2001database} and Urban100~\cite{huang2015single}. For image deraining, we evaluate using the Rain100L~\cite{yang2019joint} and Rain100H~\cite{yang2019joint} benchmarks, corresponding to light/heavy rain streaks. For evaluation on hybrid degradation, where one image contains multiple degradation types, we choose two representatives, consisting of low-resolution and noise as well as low-resolution and JPEG artifact compression, and we add noise or apply JEPG compression on the low-resolution images to synthesize second-order degraded images. For low-light image enhancement, we utilize the training and testing set of LOLv1~\cite{wei2018deep}.

\vspace{1mm}
\noindent \textbf{Evaluation Metrics.}
We use the PSNR and SSIM to evaluate the effectiveness. The PSNR/SSIM of image SR, deraining, and second-order degradation are computed on the Y channel from the YCbCr space, and we evaluate the RGB channel for denoising and low-light image enhancement. Moreover, we use trainable \#param to measure efficiency.

\vspace{1mm}
\noindent \textbf{Baseline Setup.} This work focuses on transferring pre-trained restoration models to downstream tasks under low parameter budgets. Since there is little work studying PETL on image restoration, we reproduce existing PETL approaches and compare them with the proposed AdaptIR. Specifically, we include the following representative PETL methods: \textbf{i)} VPT~\cite{jia2022visual}, where the learnable prompts are inserted as the input token of transformer layers, and we compare VPT$_{\rm{Deep}}$~\cite{jia2022visual} in experiments because of its better performance. \textbf{ii)} Adapter~\cite{houlsby2019parameter}, which introduces bottleneck structure placed after Attention and MLP. \textbf{iii)} LoRA~\cite{hu2021lora}, which adds parallel sub-networks to learn low-rank incremental matrices of query and value. \textbf{iv)} AdaptFormer~\cite{chen2022adaptformer}, which inserts a tunable module parallel to MLP. \textbf{v)} SSF~\cite{lian2022scaling}, where learnable scale and shift factors are used to modulate the frozen features. \textbf{vi)} FacT~\cite{jie2023fact}, which tensorises a ViT and then decomposes the incremental weights. We also present results of \textbf{vii)} full fine-tuning (Full-ft), and \textbf{viii)} directly applying pre-trained models to downstream tasks (Pretrain), to provide more insights. For readers unfamiliar with PETL, we have also provided a basic background introduction in~\cref{sec:background}.

\vspace{1mm}
\noindent \textbf{Implementation Details.}
We use two pre-trained transformer-based restoration models, \textit{i.e.}, IPT~\cite{chen2021pre} and EDT~\cite{li2021efficient}, as the base models to evaluate different PETL methods. We control tunable parameters by adjusting channel reduction rate $\gamma$. We use AdamW~\cite{loshchilov2017decoupled} as the optimizer and train for 500 epochs. The learning rate is initialized to 1e-4 and decayed by half at \{250,400,450,475\} epochs. All experiments are conducted on four NVIDIA 3080Ti GPUs.

\subsection{Comparison on Hybrid Degradation Tasks}
In order to obtain convincing evaluation results, it is tedious and time-consuming to observe the stability of one particular PETL method on multiple single-degradation tasks. Here, we introduce hybrid-degradation restoration. Since restoring hybrid degraded images requires a heterogeneous representation of the PETL methods, and thus the hybrid degradation is more suitable for evaluation.

In this work, we consider the second-order degradation as a representative of hybrid degradation. Specifically, we employ two different types of second-order degradations, \textit{i.e.}, the $\times$4 low-resolution and noise with $\sigma$=30 (denoted as LR4\&Noise30) as well as 
the $\times$4 low-resolution and JEPG compression with quality factor $q$=30 (denoted as LR4\&JPEG30). Moreover, we also include the classic MoE~\cite{kudugunta2021beyond,zhu2022uni,riquelme2021scaling}, which also employs the multi-branch structure but the design of each branch is the same, to give the impact of the multi-branch structure on the performance.

\begin{figure*}[!tb]
\centering
\includegraphics[width=\textwidth]{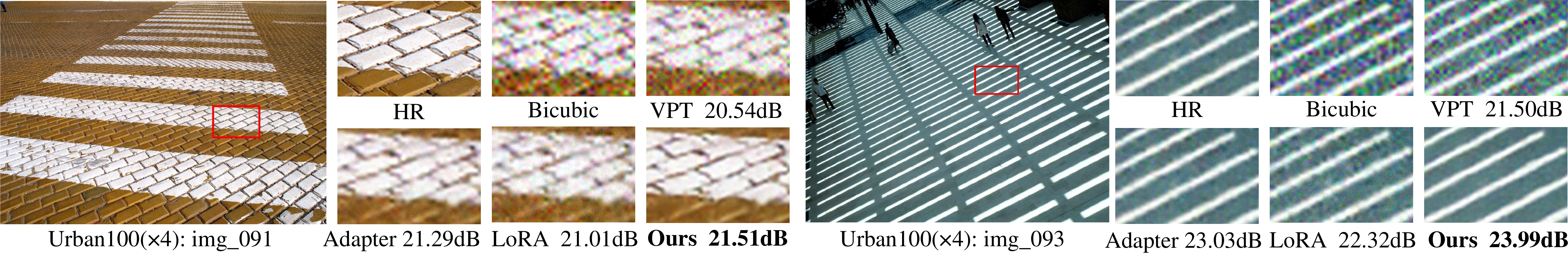}
\vspace{-4mm}
\caption{Visual comparison on hybrid degradation with LR4\&Noise30. We provide more visualization in~\cref{sec:single_task}.}
\label{fig:hybrid-degrade}
\vspace{-5mm}
\end{figure*}

\cref{table:compare-hybrid} gives the results. 
Consistent with the previous analysis in~\cref{fig:motivation}, existing PETL methods suffer severe performance drops on hybrid degradation tasks due to the difficulty of learning heterogeneous representations. Interestingly, even the simple MoE baseline which only uses the multi-branch structure outperforms the current state-of-the-art PETL methods, suggesting that multi-branch structures are promising for heterogeneity across tasks. However, since each branch of the classical MoE employs the same structure, it struggles to capture orthogonal representation bases from different branches. In contrast, our method achieves consistent state-of-the-art performance across all tasks and on all datasets. For example, our AdaptIR outperforms the state-of-the-art PETL method FacT~\cite{jie2023fact} by 1.78dB on Urban100 with LR4\&Noise30, and 0.28dB on Manga109 with LR4\&JEPG30. By orthogonally designing branches to obtain representation bases and then adaptively combining them, our AdaptIR allows for heterogeneous representations across different tasks. We also give several visual results in \cref{fig:hybrid-degrade}, and our AdaptIR can well handle complex degradation.

\begin{table*}[!t]
\centering
\caption{Quantitative comparison for hybrid-degradation restoration tasks. The best and the second best results are in \best{red} and \second{blue}.}
\label{table:compare-hybrid}
\setlength{\tabcolsep}{2.2pt}
\scalebox{0.85}{
\begin{tabular}{@{}l|c|c|cc|cc|cc|cc|cc@{}}
\toprule
\multirow{2}{*}{Method} &
  \multirow{2}{*}{Degradation} &
  \multirow{2}{*}{\#param} &
  \multicolumn{2}{c|}{\textbf{Set5}} &
  \multicolumn{2}{c|}{\textbf{Set14}} &
  \multicolumn{2}{c|}{\textbf{BSDS100}} &
  \multicolumn{2}{c|}{\textbf{Urban100}} &
  \multicolumn{2}{c}{\textbf{Manga109}} \\
              &              &      & PSNR  & SSIM   & PSNR  & SSIM   & PSNR  & SSIM   & PSNR  & SSIM   & PSNR  & SSIM   \\ \midrule
Full-ft       & LR4\&Noise30 & 119M & 27.24 & 0.7859 & 25.56 & 0.6686 & 25.02 & 0.6166 & 24.02 & 0.6967 & 26.31 & 0.8245 \\
Pretrain      & LR4\&Noise30 & -    & 19.74 & 0.3569 & 19.27 & 0.3114 & 19.09 & 0.2783 & 18.54 & 0.3254 & 19.75 & 0.3832 \\
SSF~\cite{lian2022scaling}           & LR4\&Noise30 & 373K & 25.41 & 0.6720 & 24.02 & 0.5761 & 24.06 & 0.5411 & 21.89 & 0.5514 & 23.33 & 0.6736 \\
VPT~\cite{jia2022visual}           & LR4\&Noise30 & 884K & 24.11 & 0.5570 & 22.97 & 0.4722 & 22.91 & 0.4336 & 21.20 & 0.4527 & 22.61 & 0.5570 \\
Adapter~\cite{lester2021power}       & LR4\&Noise30 & 691K & 25.60 & 0.6862 & 24.16 & 0.5856 & 24.17 & 0.5498 & 22.05 & 0.5640 & 23.61 & 0.6904 \\
LoRA~\cite{hu2021lora}          & LR4\&Noise30 & 995K & 25.19 & 0.6371 & 23.82 & 0.5405 & 23.82 & 0.5026 & 21.81 & 0.5193 & 23.30 & 0.6396 \\
Adaptfor.~\cite{chen2022adaptformer} & LR4\&Noise30 & 677K & 26.10 & 0.7138 & 24.58 & 0.6095 & 24.44 & 0.5686 & 22.52 & \second{0.5976} & {24.38} & {0.7296} \\
FacT~\cite{jie2023fact}          & LR4\&Noise30 & 537K & {25.70} & {0.6963} & {24.24} & {0.5944} & {24.25} & {0.5586} &{21.10} &{0.5727} & {23.63} & {0.6993} \\
\rowcolor[HTML]{EFEFEF} 
MoE  & LR4\&Noise30 & 667K & \second{26.35} & \second{0.7335} & \second{24.80} & \second{0.6254} & \second{24.59} & \second{0.5835}  & \second{22.77} & \best{0.6188} & \second{24.73} & \second{0.7517} \\
\rowcolor[HTML]{EFEFEF} 
Ours          & LR4\&Noise30 & 697K & \best{26.48} & \best{0.7441} & \best{24.88} & \best{0.6345} & \best{24.67} & \best{0.6279} & \best{22.88} & 0.5932 & \best{24.96} & \best{0.7625} \\ \midrule
Full-ft       & LR4\&JPEG30  & 119M & 27.21 & 0.7778 & 25.49 & 0.6563 & 25.08 & 0.6076 & 23.54 & 0.6687 & 25.48 & 0.7971 \\
Pretrain      & LR4\&JPEG30  & -    & 25.23 & 0.6702 & 24.12 & 0.5917 & 24.19 & 0.5627 & 21.74 & 0.5654 & 22.93 & 0.6732 \\
SSF~\cite{lian2022scaling}           & LR4\&JPEG30  & 373K & 26.26 & 0.7321 & 24.81 & 0.6285 & 24.71 & 0.5882 & 22.44 & 0.6085   & 23.92 & 0.7350 \\
VPT~\cite{jia2022visual}           & LR4\&JPEG30  & 884K & 26.63 & 0.7497 & 25.14 & 0.6414 & 24.89 & 0.5974 & 22.96 & 0.6377 & 24.53 & 0.7591 \\
Adapter~\cite{lester2021power}       & LR4\&JPEG30  & 691K & 26.73 & 0.7554 & 25.22 & 0.6448 &24.92 & 0.5999 & 23.09 & 0.6447  & 24.74 & 0.7677 \\
LoRA~\cite{hu2021lora}          & LR4\&JPEG30  & 995K & 26.64 & 0.7501 & 25.17 & 0.6424 & 24.91 & 0.5983 &23.02 & 0.6405  & 24.64 & 0.7619 \\
Adaptfor.~\cite{chen2022adaptformer} & LR4\&JPEG30  & 677K & {26.74} & {0.7562} & 23.08 & 0.6441 & 25.22 & 0.6447 & 24.92 & 0.5996 & 24.72 & 0.7669 \\
FacT~\cite{jie2023fact}          & LR4\&JPEG30  & 537K & {26.71} & {0.7557} & 25.22 & 0.6450 & \second{24.93} & {0.5998} &  23.08 & 0.6446 & {24.74} & {0.7681} \\
\rowcolor[HTML]{EFEFEF} 
MoE          & LR4\&JPEG30  & 667K & \second{26.80} & \second{0.7590} & \second{25.26} & \second{0.6465} & {24.04} & \second{0.6009} & \second{23.14} & \second{0.6477} & \second{24.81} & \second{0.7708} \\
\rowcolor[HTML]{EFEFEF} 
Ours          & LR4\&JPEG30  & 697K & \best{26.91} & \best{0.7646} & \best{25.34} & \best{0.6502} &\best{24.98} & \best{0.6032} & \best{23.25} & \best{0.6541}  & \best{25.02} & \best{0.7791} \\ \bottomrule
\end{tabular}%
}
\end{table*}

\subsection{Comparison with All-in-One Methods}
\vspace{-3mm}
Recently, all-in-one image restoration methods~\cite{potlapalli2023promptir,li2022all}, which learn a single restoration model for various degradations, have shown to be a promising paradigm in achieving generalized image restoration. Here, we compare our AdaptIR with these methods on both single-task and multi-task setups in ~\cref{table:compare-all-in-one}. For the single-task setting, our method achieves better PSNR results, \textit{e.g.} 0.31dB higher than PromptIR on denoising $\sigma$=50. In addition, the performance advantage of our AdaptIR still preserves the multi-task setup. For instance, our AdaptIR outperforms PromptIR by even 4.9dB PSNR and 0.016 SSIM on light rain streak removal. This is because all-in-one methods need to learn multiple degradation restoration within one model, resulting in learning difficulties, and the problem of negative transfer among different tasks~\cite{zhang2024real} can also lead to performance degradation. By contrast, the heterogeneous representation from the orthogonal design facilitates the stable performance of our AdaptIR across different degradations. As for efficiency, our AdaptIR only trains 0.7\% parameters than that of PromptIR with a fast fine-tuning process. We provide a detailed summarization and discussion about the existing multi-task restoration paradigm in \cref{sec:suppl-discuss-multi-task}.

\begin{table}[!t]
\centering
\caption{Comparison with all-in-one image restoration methods under single-task setting. The `training time' of AdaptIR refers to the downstream fine-tuning time excluding the pre-training stage.}
\label{table:compare-all-in-one}
\setlength{\tabcolsep}{3.6pt}
\scalebox{0.8}{
\begin{tabular}{@{}l|ccccccc@{}}
\toprule
Method   & task             & dataset  & \#param & training time & GPU memory & PSNR  & SSIM  \\ \midrule
AirNet~\cite{li2022all}   & light derain     & Rain100L  & 8.75M  & $\sim$48h  & $\sim$11G   & 34.90 & 0.977 \\
PromptIR~\cite{potlapalli2023promptir} & light derain     & Rain100L   & 97M     & $\sim$84h      & $\sim$128G   & 37.04 & 0.979 \\
Ours     & light derain     & Rain100L    & 697K    & $\sim$8h  & $\sim$8G   & 37.81 & 0.981 \\ \midrule
AirNet~\cite{li2022all}   & denoise $\sigma$=50 & Urban100 & 8.75M &$\sim$48h & $\sim$11G           & 28.88 & 0.871 \\
PromptIR~\cite{potlapalli2023promptir} & denoise $\sigma$=50 & Urban100  & 97M & $\sim$84h    & $\sim$128G    & 29.39 & 0.881 \\
Ours     & denoise $\sigma$=50 & Urban100    & 697K    & $\sim$8h & $\sim$8G    & 29.70 & 0.881 \\ \bottomrule
\end{tabular}%
}
\vspace{-5mm}
\end{table}

\begin{table}[!t]
\centering
\caption{Comparison with all-in-one image restoration methods under multi-task setting.}
\vspace{-1mm}
\label{tab:my-table}
\setlength{\tabcolsep}{3.6pt}
\scalebox{0.8}{
\begin{tabular}{@{}l|cccccc@{}}
\toprule
Method &
  \#param &
  \begin{tabular}[c]{@{}c@{}}GPU \\ memory\end{tabular} &
  \begin{tabular}[c]{@{}c@{}}training \\ time\end{tabular} &
  \begin{tabular}[c]{@{}c@{}}light\\ derain\end{tabular} &
  \begin{tabular}[c]{@{}c@{}}denoise\\ $\sigma$=25\end{tabular} &
  \begin{tabular}[c]{@{}c@{}}denoise\\ $\sigma$=30\end{tabular} \\ \midrule
AirNet~\cite{li2022all}   & 8.7M & $\sim$11G  & $\sim$48h & 34.90/0.967  & 31.90/0.914  & 28.68/0.861  \\
PromptIR~\cite{potlapalli2023promptir} & 97.1M   & $\sim$128G & $\sim$48h & 36.37/0.972  & 32.09/0.919  & 28.99/0.871  \\
Ours     & 697K  & $\sim$8G   & $\sim$10h & 41.27/0.988 & 32.64/0.926 & 29.16/0.875 \\ \bottomrule
\end{tabular}%
}
\vspace{-3mm}
\end{table}

\begin{figure}[!tb]  
    \begin{minipage}[t]{0.48\linewidth}
    \centering
    \captionsetup{width=0.9\linewidth}
    \caption{Fouriur analysis on outputs from LIM and FAM.}
    \vspace{-3mm}
     \includegraphics[width=0.8\linewidth]{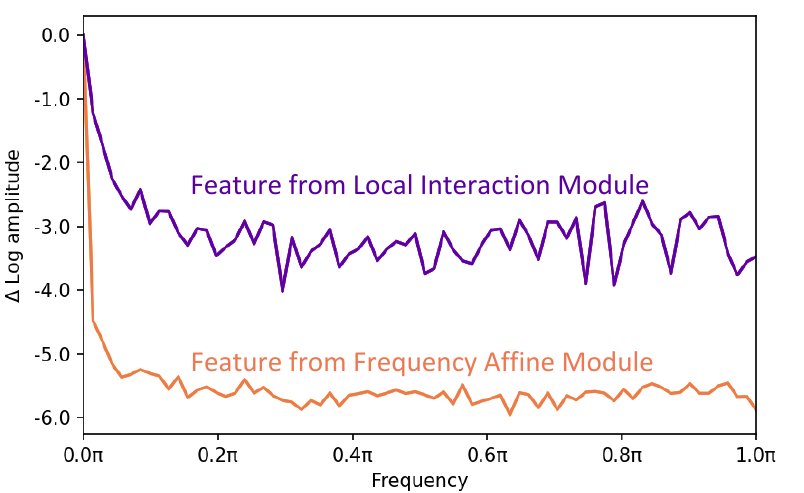}
     \label{fig:fourior}
    \end{minipage}
    \begin{minipage}[t]{0.48\linewidth}
    \vspace{0pt}
    \centering
    \captionsetup{width=0.9\linewidth}
    \caption{Channel activation visualization on outputs from CGM.}
    \vspace{-3mm}
    \includegraphics[width=0.65\columnwidth]{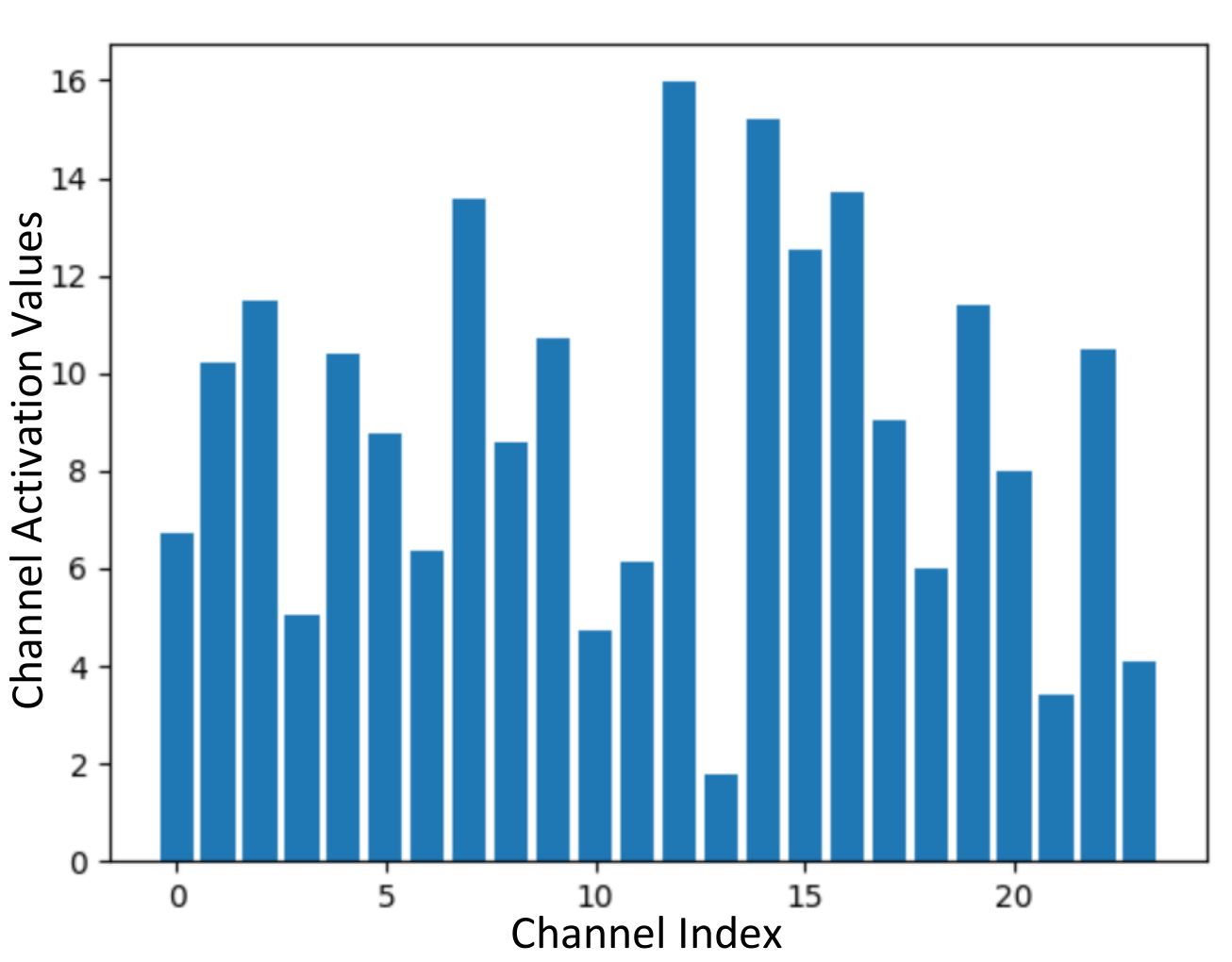}
    \label{fig:channel-activation}
    \end{minipage}
    \vspace{-7mm}
\end{figure}

\subsection{Discussion}

\noindent \textbf{Why does the Proposed Methods Work?}
Our proposed AdaptIR adopts the heterogeneous MoE structure to allow diverse representation learning. Here, we delve deep to verify whether the model design can indeed influence the learned features. For LIM and FAM, we visualize the frequency characteristics of their outputs in~\cref{fig:fourior}. It can be seen that LIM's relative log amplitude at $\pi$ is 11.02 higher than FAM, suggesting it has learned to capture high-frequency local textures. Meanwhile, more than 95\% of energy is centralized within 0.05$\pi$ for FAM, indicating it can well model low-frequency global structure. For CGM, we visualize the channel activation in~\cref{fig:channel-activation}, and find large activation differences across channels, with a large variance of 96.10, indicating that the CGM learns to select degradation-specific channels.

\begin{figure}[!tb]  
    \begin{minipage}[t]{0.48\linewidth}
        \centering
        \captionsetup{width=0.96\linewidth}
        \captionof{table}{Comparison on generalization ability with more pretrained base model.}
\label{table:compare-edt}
        \vspace{1mm}
        \setlength{\tabcolsep}{1.8pt}
        \scalebox{0.74}{
        \begin{tabular}{@{}l|c|ccccc@{}}
        \toprule
         &
           &
           &
           &
           &
           &
           \\
        \multirow{-2}{*}{Method} &
          \multirow{-2}{*}{\#param} &
          \multirow{-2}{*}{{Set5}} &
          \multirow{-2}{*}{{Set14}} &
          \multirow{-2}{*}{{\begin{tabular}[c]{@{}c@{}}BSDS\\ 100\end{tabular}}} &
          \multirow{-2}{*}{{\begin{tabular}[c]{@{}c@{}}Urban\\ 100\end{tabular}}} &
          \multirow{-2}{*}{{\begin{tabular}[c]{@{}c@{}}Manga\\ 109\end{tabular}}} \\ \midrule
        Full-ft &
          11.6M &
          27.32 &
          25.60 &
          25.03&
          24.10 &
          26.42 \\
        Pretrain &
          - &
          19.29 &
          18.45 &
          18.27 &
          17.92 &
          19.25 \\
        SSF~\cite{lian2022scaling} &
          117K &
          {\color[HTML]{333333} 26.92} &
          {\color[HTML]{333333} 25.24} &
          {\color[HTML]{333333} 24.83} &
          {\color[HTML]{333333} 23.41} &
          {\color[HTML]{333333} 25.77} \\
        VPT~\cite{jia2022visual} &
          311K &
          24.19 &
          22.91 &
          22.81 &
           21.12&
          22.49 \\
        Adapter~\cite{houlsby2019parameter} &
          194K &
          {\color[HTML]{333333} 26.92} &
          {\color[HTML]{333333} 25.27} &
          {\color[HTML]{333333} 24.81} &
          {\color[HTML]{333333} 23.48} &
          {\color[HTML]{333333} 25.84} \\
        LoRA~\cite{hu2021lora} &
          259K &
          {\color[HTML]{000000} 26.91} &
          {\color[HTML]{000000} 25.25} &
          {\color[HTML]{000000} 24.80} &
          {\color[HTML]{000000} 23.46} &
          25.81 \\
        AdaptFor.~\cite{chen2022adaptformer} &
          162K &
          {\color[HTML]{0000FF} 26.99} &
          {\color[HTML]{0000FF} 25.31} &
          {\color[HTML]{0000FF} 24.85} &
          {\color[HTML]{0000FF} 23.59} &
          {\color[HTML]{0000FF} 25.95} \\
        FacT~\cite{jie2023fact} &
          174K &
          {\color[HTML]{333333} 26.89} &
          {\color[HTML]{333333} 25.25} &
          {\color[HTML]{333333} 24.81} &
          {\color[HTML]{333333} 23.43} &
          {\color[HTML]{333333} 25.78} \\
        Ours &
          173K &
          {\color[HTML]{FF2121} 27.04} &
          {\color[HTML]{FF2121} 25.34} &
          {\color[HTML]{FF2121} 24.87} &
          {\color[HTML]{FF2121} 23.60} &
          {\color[HTML]{FF2121} 25.97} \\ \bottomrule
        \end{tabular}%
        }
    \end{minipage}
    \begin{minipage}[t]{0.48\linewidth}
    \vspace{0pt}
    \centering
    \captionsetup{width=0.9\linewidth}
    \caption{Scalability comparison with different PETL methods.}
    \includegraphics[width=1\columnwidth]{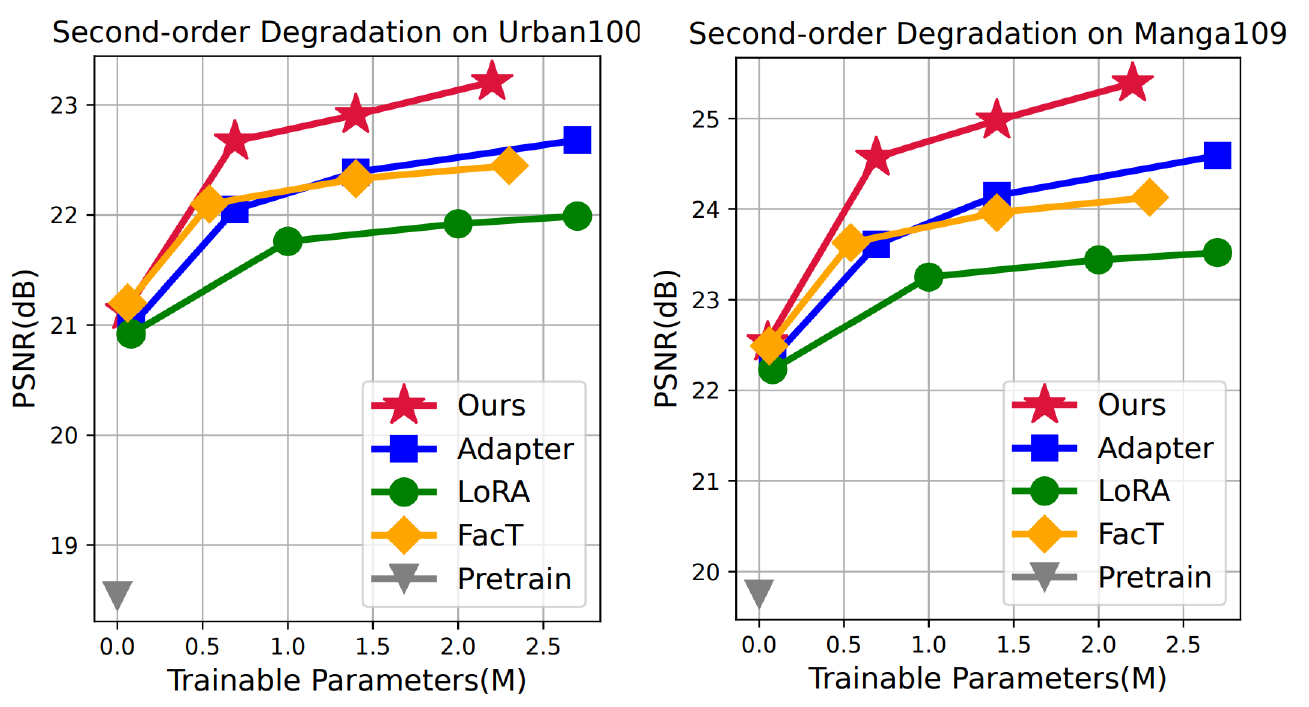}
    \label{fig:scalability}
    \end{minipage}
    \vspace{-9mm}
\end{figure}

\noindent \textbf{Scaling Trainable Parameters.}
We compare the performance of different PETL methods under varying parameter budgets. We use the hybrid degradation LR4\&Noise30 in this setup. \cref{fig:scalability} shows the results. It can be seen that the proposed method surpasses other strong baselines across various parameter settings, demonstrating the strong scalability of the proposed method.

\vspace{1mm}
\noindent \textbf{How About the Performance on Other Pre-trained Models?}
The above experiments employ IPT~\cite{chen2021pre} as the base model. In order to verify the generalization of the proposed method, we further adopt another pre-trained image restoration model EDT~\cite{li2021efficient} as the frozen base model. \cref{table:compare-edt} represents the results. It can be seen that the proposed method maintains state-of-the-art performance by tuning only 1.5\% parameters. More experiments with EDT can be seen in~\cref{sec:more-edt}.

\begin{table}[!tb]
\centering
\caption{Ablation of different design choices on PSNR(dB). `Baseline' refers to the setting of depth-separable projection in LIM and FAM, as well as the channel-spatial orthogonal modeling.}
\label{table:ablation-efficent-design}
\setlength{\tabcolsep}{2.0pt}
\scalebox{0.82}{
\begin{tabular}{@{}l|c|ccc@{}}
\toprule
Settings                     & \#param & Set5  & Set14 & Urban100 \\ \midrule
(0)Baseline                  & 697K    & 26.48 & 24.88 & 22.88    \\
(1)+w/o adaptive feature ensemble & 692K & 24.26 & 24.88 & 22.82 \\ 
(2)+w/o depth-separable in LIM      & 718K    & 25.67 & 24.22 & 22.10    \\
(3)+w/o depth-separable in FAM     & 728K    & 25.73 & 24.28 & 22.17    \\
(4)+w/o CGM\&w/o depth-separable & 743K    & 24.26 & 23.15 & 21.36    \\ 
\bottomrule
\end{tabular}%
}
\end{table}

\subsection{Ablation Study}

\noindent \textbf{Parameter Efficient Designs.}
In this work, we introduce several techniques to achieve orthogonal representation learning. Here, we ablate to study the impact of these choices. The results, presented in \cref{table:ablation-efficent-design}, indicate that (1) the adaptive feature ensemble can assemble representations according to specific degradation, without which will cause a performance drop. In addition, (2)\&(3) removing the depth-separable design in LIM or FAM will conflict with the channel modeling in CGM, and lead to sub-optimal results. Further, (4) removing the CGM branch while allowing full-channel interaction in other branches results in poor performance, which we attribute to the learning difficulty of modeling channel and spatial simultaneously.

\noindent \textbf{Ablation for Components.}
In the proposed AdaptIR, three parallel branches are developed to learn orthogonal bases. We ablate to discern the roles of different branches. As shown in \cref{table:ablation-component}, separate utilization of one or two branches only yields sub-optimal results owing to the insufficient representation. And the combination of the three branches achieves the best results.

\begin{figure}[!tb]  
    \begin{minipage}[t]{0.48\linewidth}
        \centering
        \captionsetup{width=0.94\linewidth}
        \captionof{table}{Ablation experiments of different components on PSNR(dB).}
        \label{table:ablation-component}
        \setlength{\tabcolsep}{1.8pt}
        \scalebox{0.799}{
        \begin{tabular}{@{}ccc|c|ccc@{}}
        \toprule
        LIM       & FAM       & CGM       & \#param & Set5  & Set14 & Urban100 \\ \midrule
                  &           & \ding{52} & 680K    & 23.64 & 22.66 & 20.97    \\
                  & \ding{52} & \ding{52} & 682K    & 25.52 & 24.17 & 22.05    \\
        \ding{52} & \ding{52} &           & 678K    & 26.11 & 24.60 & 22.52    \\
        \ding{52} & \ding{52} & \ding{52} & 697K    & 26.48 & 24.88 & 22.88    \\ \bottomrule
        \end{tabular}%
        }
    \end{minipage}
    \begin{minipage}[t]{0.48\linewidth}
    \vspace{0pt}
    \centering
    \captionsetup{width=0.94\linewidth}
    \captionof{table}{Ablation for different insertion positions and forms on PSNR(dB).}
    \label{table:ablation-insert-place}
    \setlength{\tabcolsep}{1.8pt}
    \scalebox{0.799}{
    \begin{tabular}{@{}l|c|ccc@{}}
    \toprule
    position  & form       & Set5  & Set14 & Urban100 \\ \midrule
    MLP       & parallel   & 26.48 & 24.88 & 22.88    \\
    Attention & parallel   & 26.28 & 24.70 & 22.59    \\
    MLP       & sequential & 26.35 & 24.77 & 22.67    \\
    Attention & sequential & 25.60 & 24.19 & 22.07    \\ \bottomrule
    \end{tabular}%
    }
    \end{minipage}
    \vspace{-5mm}
\end{figure}

\noindent \textbf{Insertion Position and Form.}
There are various options for both the insertion location and form of our AdapIR. The impact of these choices is shown in \cref{table:ablation-insert-place}. It can be seen that inserting AdaptIR into MLP achieves better performance under both parallel and sequential forms. This is because there is a certain dependency between the well-trained MLP and attention, and insertion into the middle of them
will damage this relationship. Moreover, the parallel insertion form performs better than its sequential counterpart. We argue that parallel form can preserve the knowledge of frozen features through summation, thus reducing the learning difficulty.

\section{Conclusion}
\vspace{-2mm}
In this work, we explore for the first time the potential of parameter-efficient adaptation to improve the generalization of image restoration models. We observe that current PETL methods struggle to generalize to multiple single-degradation tasks and suffer from performance degradation on hybrid-degradation tasks. We identify that this issue arises from the misalignment between the degradation-required representation and the homogeneity in current PETL methods. Based on this observation, we propose AdaptIR, a heterogeneous Mixture-of-Experts (MoE) to learn local spatial, global spatial, and channel orthogonal bases under low parameter budgets, followed by the adaptive feature ensemble to dynamically fuse these bases for degradation-specific representation. Extensive experiments validate our AdaptIR as a versatile and powerful adaptation solution.

\clearpage

\section*{Acknowledgements}
\vspace{-3mm}
This work is supported in part by the National Natural Science Foundation of China, under Grant (62302309,62171248), Shenzhen Science and Technology Program (JCYJ20220818101014030, JCYJ20220818101012025), and the PCNL KEY project (PCL2023AS6-1).

{
  \small
  \bibliographystyle{unsrt}
  \bibliography{neurips_2024}
}
\clearpage
\appendix
\section*{\Large{Appendix}}
\label{sec:Appendix}

\section{Basic Background of Parameter Efficient Transfer Learning} 
\label{sec:background}

\begin{figure*}[h]
    \centering
    \includegraphics[width=0.98\linewidth]{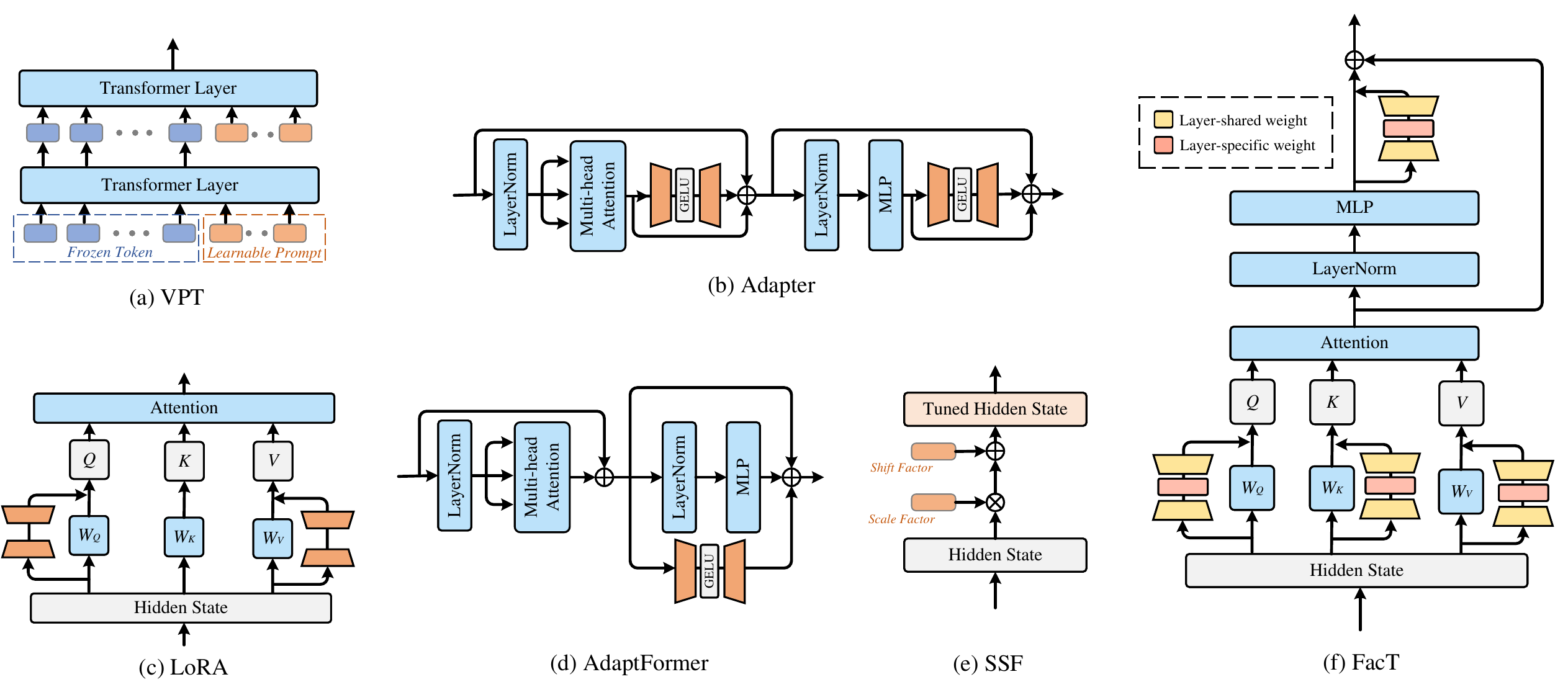}
    \caption{Illustrations of exsiting state-of-the-art PETL baselines.}
    \label{fig:baseline}
\end{figure*}

Since very little work has been done to study PETL in low-level vision, we re-implement the current state-of-the-art PETL methods in this work, such as VPT~\cite{jia2022visual}, Adapter~\cite{houlsby2019parameter}, LoRA~\cite{hu2021lora}, AdaptFormer~\cite{chen2022adaptformer}, SSF~\cite{lian2022scaling}, and FacT~\cite{jie2023fact}. In this part, we review these methods and provide detailed implementation details for reproduction. \cref{fig:baseline} gives an illustration of these baseline methods.

\begin{itemize}
    \item VPT~\cite{jia2022visual}, shown in ~\cref{fig:baseline}(a), prepends learnable prompt tokens in the input of one transformer layer~\cite{vaswani2017attention}. In ~\cite{jia2022visual}, there are two versions of VPT, \textit{i.e.}, VPT$_{\rm{Shallow}}$ and VPT$_{\rm{Deep}}$. To obtain a good performance, we use VPT$_{\rm{Deep}}$ which inserts new prompts at each transformer layer, as our default settings. 
    \item Adapter~\cite{houlsby2019parameter}, shown in \cref{fig:baseline}(b), is a bottleneck structure with an intermediate GELU activation function. Following the vanilla Adapter design~\cite{houlsby2019parameter}, we insert the Adapter both after the Multi-head Self-Attention (MSA) and Multi-Layer Perceptron (MLP).
    \item LoRA~\cite{hu2021lora}, shown in \cref{fig:baseline}(c) use the multiplication of two low-rank to approximate the incremental matrices in projection layers of Query and Value.
    \item AdaptFormer~\cite{chen2022adaptformer}, shown in \cref{fig:baseline}(d), is similar in model architecture with Adapter, but has different insert position and form. In ~\cite{chen2022adaptformer}, the AdaptFormer is placed before the second LayerNorm layer and adopts a parallel insertion form.
    \item SSF~\cite{lian2022scaling}, shown in \cref{fig:baseline}(e), utilize learnable scale and shift factors to modulate frozen features. Based on the settings in~\cite{lian2022scaling}, we place the SSF layer behind all the attention QKV projection, the LayerNorm, and the MLP layers.
    \item FacT~\cite{jie2023fact}, shown in \cref{fig:baseline}(f), tensorises Vision Transformer~\cite{dosovitskiy2020image} and introduces a low-rank approximation to the incremental matrix similar to LoRA. Different from LoRA, FacT sets the up-projection and down-projection to be shared across layers while setting the projection in the low-rank space to be layer-specific. There are two versions of FacT, namely $\rm{FacT_{\rm{TT}}}$ and $\rm{FacT_{\rm{TK}}}$~\cite{jie2023fact}, we use $\rm{FacT_{TT}}$ in this work because of its good performance. Following ~\cite{jie2023fact}, we introduce the FacT layer in attention QKV projection as well as MLP layers.
     
\end{itemize}

\section{Discussion with Prompt-based Methods}
In this section, we briefly discuss the difference between our AdaptIR with other prompt-based methods. To the best of our knowledge, only the PromtGIP~\cite{liu2023pmtgip} shows the zero-shot ability when facing unseen degradations. And the other ProRes~\cite{ma2023prores}, PromptIR~\cite{potlapalli2023promptir}, and PromptRestorer~\cite{wang2023promptrestorer} can only handle degradations which have been seen during training, which means they still need additional fine-tuning for task generalization. The advantages of our AdaptIR compared to these prompt-based approaches are two-fold. As for efficiency, PromptIR, ProRes, and PromptRestorer all need full fine-tuning for adapting to new tasks, \textit{e.g.}, PromptIR needs 7-day 8$\times$3090 GPUs for full fine-tuning, while our AdaptIR needs only 8h 1$\times$3090 GPU. As for performance, since these methods need to learn multiple degradations within one model, it is inevitable to suffer the problem of negative transfer, which impairs performance. We give a thorough comparison in ~\cref{table:suppl-compare-prompt-methods}.

\begin{table}[!tb]
\centering
\caption{Comparison with prompt-based restoration methods.}
\label{table:suppl-compare-prompt-methods}
\setlength{\tabcolsep}{2pt}
\scalebox{0.85}{
\begin{tabular}{lccccc}
\toprule
\textbf{Methods} & \textbf{Type} & \textbf{Fast Adaptation} & \textbf{Adaptation Cost} & \textbf{PSNR on denoising} & \textbf{PSNR on deraining} \\ \midrule
PromptGIP~\cite{liu2023pmtgip} & Prompt-based & Yes & zero-shot & 26.22 & 25.46 \\ \midrule
ProRes~\cite{ma2023prores} & Prompt-based & No & 8x3090GPUs & Not open-source & Not open-source \\ \midrule
PromptIR~\cite{potlapalli2023promptir}& Prompt-based & No & 7-days 8x3090GPUs & 29.39 & 37.04 \\ \midrule
PromptRestorer~\cite{wang2023promptrestorer} & Prompt-based & No & 8x3090GPUs & Not open-source & Not open-source \\ \midrule
Ours & PETL-based & Yes & 8h 1x3090 & 29.70 & 37.81 \\ \bottomrule
\end{tabular}
}
\end{table}

\section{Discussion on Multi-task Restoration Paradigms.}
\label{sec:suppl-discuss-multi-task}
In this section, we revisit existing paradigms for dealing with multi-task restoration problems, in which multiple degradations need to be handled. Let denote the number of degradation types, \textit{i.e.}, downstream tasks as $N$, then we can summarize existing paradigm into the following three categories:
\begin{enumerate}
    \item ``N for N'': training N task-specific models for N downstream tasks, such as Restormer, MPRNet.
    \item ``1 for N'': training 1 all-in-one model for N downstream tasks, such as PromptIR~\cite{potlapalli2023promptir}, AirNet~\cite{li2022all}.
    \item ``(1+N) for N'': using 1 task-shared pre-trained weights, and N task-specific lightweight modules.
\end{enumerate}

Early image restoration techniques predominantly employed the first strategy, which trains $N$ different models to handle multiple degradations. Although this strategy can handle multiple degradations, it usually requires training and storing $N$ copies for each task. The recent all-in-one methods train one model for multiple degradations, although reducing the model copy to one to improve the efficiency, this approach usually suffers from performance degradation due to the multi-task learning difficulties and the negative transfer learning problem. In this work, the proposed AdaptIR is the first paradigm categorized in the third category, which trains a shared pre-trained backbone as well as $N$ task-specific lightweight modules. This paradigm can be seen as a compromise between effectiveness and efficiency. However, given that the task-specific modules are very lightweight, we believe that the advantages of this paradigm outweigh the disadvantages.

\section{Further Explanation of the Heterogeneous Representation.}
\begin{figure*}[h]
    \centering
    \includegraphics[width=0.98\linewidth]{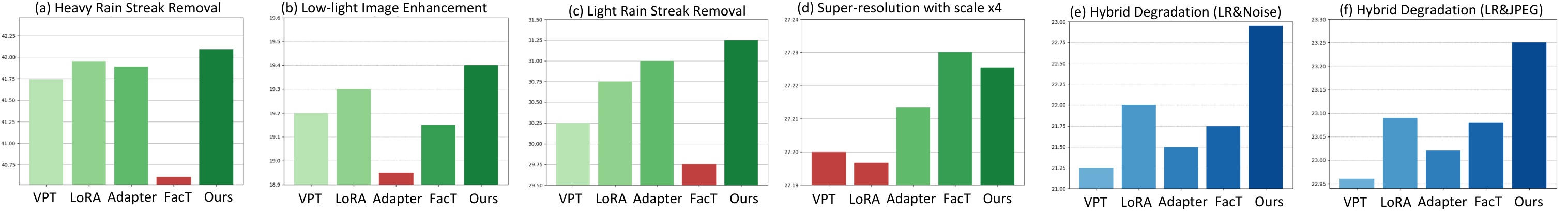}
    \caption{More evidence on the unstable performance of previous PETL methods across different single-degradation types, and the unfavorable performance under hybrid degradation.}
    \label{fig:suppl-more-evidence-on-performace}
\end{figure*}

\begin{figure*}[h]
    \centering
    \includegraphics[width=0.98\linewidth]{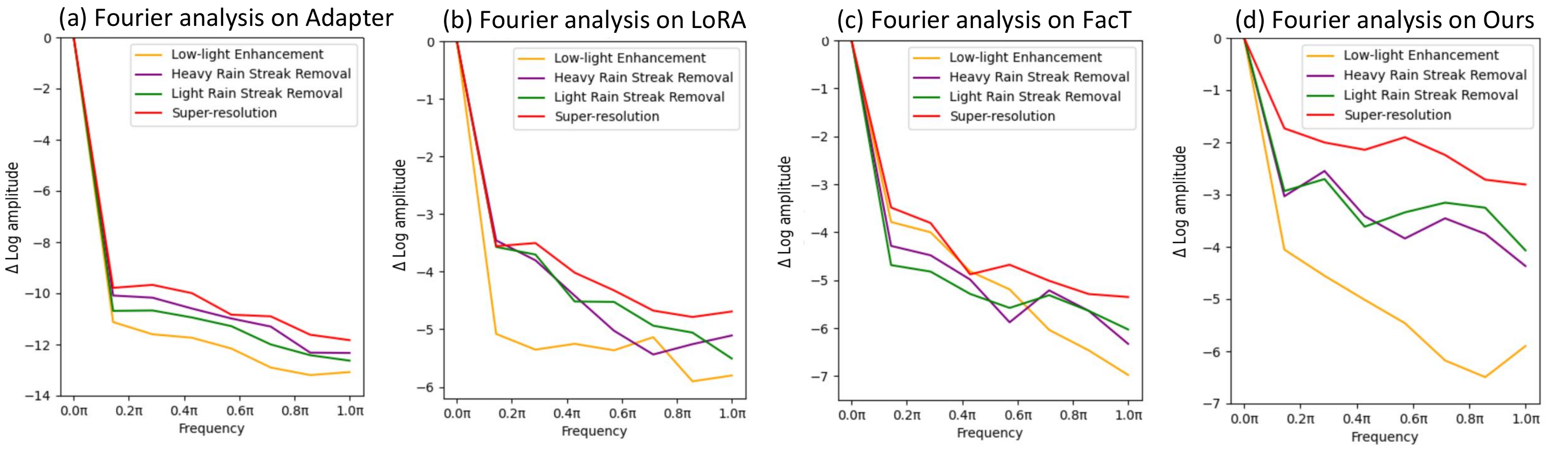}
    \caption{More evidence that shows previous PETL methods struggle to learn distinguishable features across different degradation types, i.e., homogeneous representation. In contrast, our AdaptIR can learn heterogeneous representations for different degradations.}
    \label{fig:suppl-more-fourior-analysis}
\end{figure*}

In this work, we pay attention to the learning of the \textit{heterogeneous representation}. Here, we articulate it to make it more clear about this term. The heterogeneous representation in this paper represents the learning of discriminative features across different degradation types. The term representation here is instantiated as the Fourier curve in ~\cref{fig:motivation} in the main paper. Previous approaches tend to produce similar representations across various degradations. As common knowledge, restoring different degradations requires different representations, \textit{e.g.}, SR needs a high-pass filter network while denoising needs low-pass. As a result, if the representation needed by current degradation matches the specific representation of the existing PETL method, it works. If not, it leads to unstable performance. To demonstrate the generality of the problem regarding the unstable performance and the homogeneous representation under different degradations, we provide more evidence in ~\cref{fig:suppl-more-evidence-on-performace} and ~\cref{fig:suppl-more-fourior-analysis}.

\section{More Experiment Results}
\label{sec:more-exp}

\subsection{Comparison on Single-degradation Tasks}
\label{sec:single_task}

\begin{figure*}[!t]
\centering
\includegraphics[width=1\textwidth]{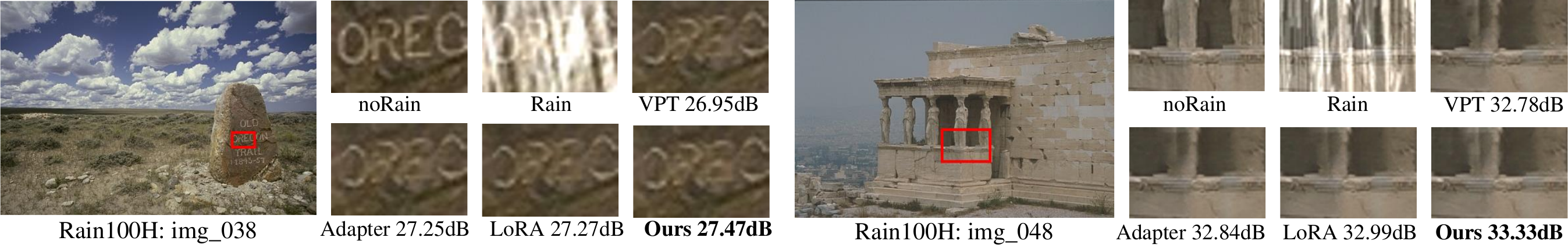}
\vspace{-2mm}
\caption{Visual comparison of heavy rain streak removal on samples from Rain100H~\cite{yang2019joint} dataset.}
\label{fig:rain100H}
\end{figure*}

\begin{figure*}[!tb]
\centering
\includegraphics[width=1\textwidth]{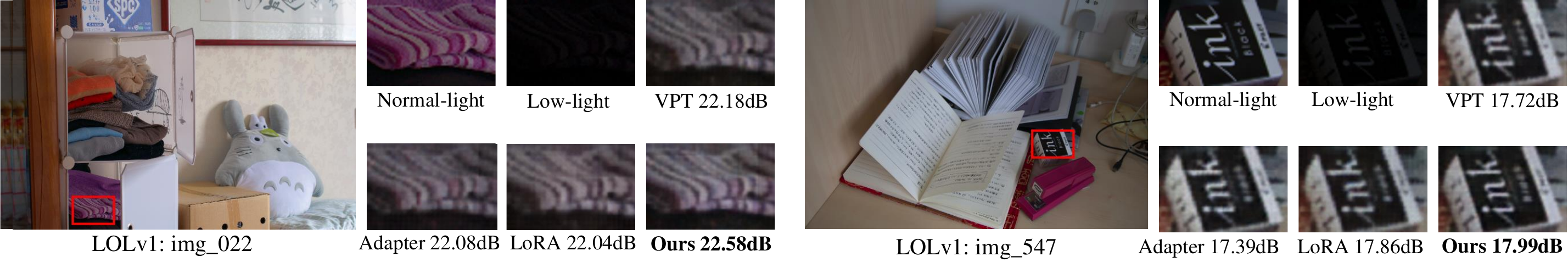}
\vspace{-2mm}
\caption{Visual comparison of low-light image enhancement on samples from LOLv1~\cite{wei2018deep} dataset.}
\label{fig:low-light}
\end{figure*}

Current PETL methods struggle to achieve stable performance due to homogeneous frequency characteristics and suffer performance degradation when not well aligned with the frequencies required by a specific degradation. Therefore, it needs multiple single-degradation tasks to obtain convincing evaluation results. Here, we give results of  single-degradation tasks, including \textbf{image super-resolution} in~\cref{table:compare-sr}, \textbf{color image denoising} in~\cref{table:compare-dn+derain}, \textbf{light deraining} in~\cref{table:rainL}, \textbf{heavy deraining} in~\cref{table:rainH}, and \textbf{low-light enhancement} in~\cref{table:low-light}. It can be seen that our method maintains a stable best performance on most single-degradation tasks, with the second-best method varying across tasks. For example, the recent state-of-the-art methods FacT~\cite{jie2023fact} obtains comparable performance with our AdaptIR, however, it suffers significant performance degradation on the subsequent light and heavy rain streak removal tasks. Another example can also be seen in LoRA \cite{hu2021lora}, which performs the second best in heavy deraining but struggles with low-light image enhancement tasks. In contrast, our method is more stable, achieving consistent best performance across these single-degradation tasks.
We also provide quantitative comparisons on the single-degradation restoration tasks in~\cref{fig:rain100H} and ~\cref{fig:low-light}.

\subsection{Additional Results on Hybrid Degradation with EDT}
\label{sec:more-edt}
In order to demonstrate the generalizability of our AdaptIR, we choose IPT~\cite{chen2021pre} and EDT~\cite{li2021efficient} as pre-trained base models to evaluate the performance of different PETL methods. Due to the page limit, we mainly present the results of IPT in the main paper. Here, we give more experimental results with EDT. The results on second-order degradation LR4\&JPEG30 with EDT are shown in \cref{table:more-edt}. It can be seen that our method continues to achieve state-of-the-art performance by a significant margin. For example, our AdaptIR outperforms the second-best method FacT~\cite{jie2023fact} by up to 0.15dB PSNR while using fewer tunable parameters. The results with EDT as the base model demonstrate the robustness of the proposed method.

\begin{figure}[!tb]  
    \begin{minipage}[t]{0.48\linewidth}
        \centering
        \captionsetup{width=0.94\linewidth}
        \captionof{table}{Quantitative comparison for $\times 4$ {image SR} on PSNR(dB). We compare the 
        \#param when the performance is the same.}
        \label{table:compare-sr}
        \setlength{\tabcolsep}{1.8pt}
        \scalebox{0.799}{
        \begin{tabular}{@{}l|c|ccccc@{}}
        \toprule
         &
           &
           &
           &
           &
           &
           \\
        \multirow{-2}{*}{Method} &
          \multirow{-2}{*}{\#param} &
          \multirow{-2}{*}{\textbf{Set5}} &
          \multirow{-2}{*}{\textbf{Set14}} &
          \multirow{-2}{*}{\textbf{\begin{tabular}[c]{@{}c@{}}BSDS\\ 100\end{tabular}}} &
          \multirow{-2}{*}{\textbf{\begin{tabular}[c]{@{}c@{}}Urban\\ 100\end{tabular}}} &
          \multirow{-2}{*}{\textbf{\begin{tabular}[c]{@{}c@{}}Manga\\ 109\end{tabular}}} \\ \midrule
        RCAN~\cite{zhang2018image} &
          15.6M &
          32.63 &
          28.87 &
          27.77 &
          26.82 &
          31.22 \\
        SAN~\cite{dai2019second} &
          15.9M &
          32.64 &
          28.92 &
          27.78 &
          26.79 &
          31.18 \\
        SwinIR~\cite{liang2021swinir} &
          11.9M &
          32.74 &
          29.06 &
          27.89 &
          27.37 &
          31.93 \\ \midrule
        Full-ft &
          119M &
          32.66 &
          29.03 &
          27.82 &
          27.31 &
          31.64 \\
        pretrain &
          - &
          32.58 &
          28.97 &
          27.79 &
          27.18 &
          31.41 \\
        VPT~\cite{jia2022visual} &
          884K &
          32.71 &
          29.02 &
          27.82 &
          27.20 &
          31.65 \\
        Adapter~\cite{houlsby2019parameter} &
          691K &
          32.70 &
          29.03 &
          27.82 &
          27.21 &
          31.68 \\
        LoRA~\cite{hu2021lora} &
          995K &
          32.70 &
          29.03 &
          27.82 &
          27.20 &
          31.68 \\
        Adaptfor.~\cite{chen2022adaptformer} &
          677K &
          32.70 &
           29.03 &
          27.82 &
          27.21 &
           31.68 \\
        SSF~\cite{lian2022scaling} &
          373K &
          29.56 &
          26.84 &
          26.50 &
          23.78 &
          26.02 \\
        FacT~\cite{jie2023fact}    & 537K                      & {\color[HTML]{0000FF}32.71}                        & {\color[HTML]{0000FF}29.03}                            &{\color[HTML]{0000FF}27.82}                                                                     &  {\color[HTML]{FF2121}27.23}                                                                    & {\color[HTML]{0000FF}31.70}                                                                         \\
Ours                          & 370K                      & {\color[HTML]{FF2121}32.71}    & {\color[HTML]{FF2121}29.04}                            & {\color[HTML]{FF2121} 27.82}                                                  &  {\color[HTML]{0000FF}27.22}                                                  & {\color[HTML]{FF2121} 31.70}       
         \\ \bottomrule
        \end{tabular}%
        }
    \end{minipage}
    \begin{minipage}[t]{0.48\linewidth}
    \vspace{0pt}
    \centering
    \captionsetup{width=0.94\linewidth}
    \captionof{table}{Quantitative comparison for {color image denoising} on PSNR(dB). We compare the \#param when PSNR is the same.}
    \label{table:compare-dn+derain}
    \setlength{\tabcolsep}{1.8pt}
    \scalebox{0.799}{
    \begin{tabular}{@{}l|c|cc|cc@{}}
    \toprule
    \multirow{2}{*}{Method} & \multirow{2}{*}{\#param} & \multicolumn{2}{c|}{\textbf{CBSD68}} & \multicolumn{2}{c}{\textbf{Urban100}} \\
                            &                   & $\sigma$=30     & $\sigma$=50    & $\sigma$=30      & $\sigma$=50     \\ \midrule
    FFDNet~\cite{zhang2018ffdnet}                  & 0.8M                        & 30.31        & 27.96       & 30.53         & 28.05        \\
    RDN~\cite{zhang2018residual}                     & 15.6M                        & 30.67        & 28.31       & 31.69         & 29.29        \\
    SwinIR ~\cite{liang2021swinir}                 & 11.9M                    & -            & 28.56       & -             & 29.82        \\ \midrule
    Full-ft                 & 119M                     & 30.75        & 28.39       & 32.01         & 29.72        \\
    Pretrain                & -                        & 30.73        & 28.36       & 31.94         & 29.68        \\
    VPT~\cite{jia2022visual}                     & 884K                     & 30.74        & 28.37       & 31.97         & 29.68        \\
    Adapter~\cite{houlsby2019parameter}                 & 691K                     & 30.74        & 28.37       & 31.97         & 29.69        \\
    LoRA~\cite{hu2021lora}                    & 995K                     & 30.75        & 28.38       & 31.98         & 29.70        \\
    AdaptFor.~\cite{chen2022adaptformer}               & 677K                     & 30.73        & 28.37       & 31.97         & 29.68        \\
    SSF~\cite{lian2022scaling}                     & 373K                     & 30.07        & 27.64       & 29.79         & 27.01        \\
FacT~\cite{jie2023fact}      & 537K         & {\color[HTML]{FF2121}30.75}                        & {\color[HTML]{0000FF}28.38}      & {\color[HTML]{0000FF}31.98 }             & {\color[HTML]{FF2121}29.70}                 \\
Ours & 515K &  {\color[HTML]{0000FF}30.74} & {\color[HTML]{FF2121} 28.38} & {\color[HTML]{FF2121} 31.98} & {\color[HTML]{0000FF} 29.69}    
    \\ \bottomrule
    \end{tabular}
    }
    \end{minipage}
\end{figure}

\begin{figure}[!tb]  
    \begin{minipage}[t]{0.48\linewidth}
        \centering
        \captionsetup{width=0.96\linewidth}
        \captionof{table}{Quantitative comparison for light rain streak removal on Rain100L dataset.}
        \label{table:rainL}
        \setlength{\tabcolsep}{1.8pt}
        \scalebox{0.799}{
        \begin{tabular}{@{}l|ccc@{}}
        \toprule
        Method    & \#param & PSNR  & SSIM   \\ \midrule
        Full-ft  & 119M      & 42.14 & 0.9905 \\
        Pretrain  & -       & 17.30 & 0.5488 \\
        VPT~\cite{jia2022visual}     & 884K    & 41.74 & 0.9896 \\
        Adapter~\cite{houlsby2019parameter} & 691K    & {\color[HTML]{0000FF}41.95}&0.9900 \\
        LoRA~\cite{hu2021lora} & 995K  & 41.89 & 0.9898 \\
        AdaptFor.~\cite{chen2022adaptformer} & 677K    &41.90& 0.8992  \\
        FacT~\cite{jie2023fact}      & 537K    &40.61& \best{0.9984}  \\
        Ours      & 697K    &    \best{42.09} & {\color[HTML]{0000FF}0.9902} \\ \bottomrule
        \end{tabular}%
        }
    \end{minipage}
    \begin{minipage}[t]{0.48\linewidth}
    \vspace{0pt}
    \centering
        \captionsetup{width=0.96\linewidth}
        \captionof{table}{Quantitative comparison for heavy rain streak removal on Rain100H dataset.}
        \label{table:rainH}
        \setlength{\tabcolsep}{1.8pt}
        \scalebox{0.799}{
        \begin{tabular}{@{}l|ccc@{}}
        \toprule
        Method    & \#param & PSNR  & SSIM   \\ \midrule
        Full-ft  & 119M      & 32.23 & 0.9202 \\
        Pretrain  & -       & 17.30 & 0.5488 \\
        VPT~\cite{jia2022visual}     & 884K    & 30.87 & 0.8967 \\
        Adapter~\cite{houlsby2019parameter} & 691K    & 30.99 & 0.8971 \\
        LoRA~\cite{hu2021lora} & 995K  & {\color[HTML]{0000FF}31.16} & {\color[HTML]{0000FF}0.9002} \\
        AdaptFor.~\cite{chen2022adaptformer} & 677K    & 31.10 & 0.8992 \\
        FacT~\cite{jie2023fact}      & 537K    & 29.70 & 0.8824 \\
        Ours      & 697K    & \best{31.23} & \best{0.9016} \\ \bottomrule
        \end{tabular}%
        }
    \end{minipage}
\end{figure}

\begin{table}[!tb]
\centering
\caption{Quantitative comparison for low-light image enhancement with LOLv1 dataset.}
\label{table:low-light}
\setlength{\tabcolsep}{2pt}
\scalebox{0.85}{
\begin{tabular}{@{}l|cccccccccc@{}}
\toprule
Metric  & Pretrain &\makecell{UFormer \\ \cite{wang2022uformer}}&\makecell{RetinexNet \\ \cite{Chen2018Retinex}}& \makecell{FIDE \\ \cite{xu2020learning}} &\makecell{VPT\\ \cite{jia2022visual}}  & \makecell{Adapter \\  \cite{houlsby2019parameter}}& \makecell{LoRA \\ \cite{loshchilov2017decoupled} } & \makecell{AdaptFor. \\ \cite{chen2022adaptformer} }& \makecell{FacT\\ \cite{jie2023fact}} & \makecell{ AdaptIR \\ (ours)} \\ \midrule
\#param & -     &-&-&-   & 884K & 691K    & 995K & 677K        & 537K & 697K \\
PSNR & 7.64&16.36  &16.77&18.27& 19.28  & 19.22  & 18.94  & {\color[HTML]{0000FF}19.40}  & 19.06 & \best{19.46}  \\
SSIM & 0.2547&0.771& 0.560 &0.665& 0.7198 & 0.7293 & 0.7197 & {\color[HTML]{0000FF}0.7352} & 0.7147 & \best{0.7441} \\ \bottomrule
\end{tabular}%
}
\end{table}

\begin{table}[tb]
\centering
\caption{Quantitative comparison for second-order degradation with LR4\&JPEG30 using  EDT as pre-trained restoration models. The best results are \textbf{bolded}.}
\setlength{\tabcolsep}{1.8pt}
\scalebox{0.98}{
\begin{tabular}{l|c|ccccc}
\toprule
\multirow{2}{*}{Method} &
  \multirow{2}{*}{\#param} &
  \textbf{Set5} &
  \textbf{Set14} &
  \textbf{BSDS100} &
  \textbf{Urban100} &
  \textbf{Manga109} \\
              &       & PSNR/SSIM             & PSNR/SSIM             & PSNR/SSIM             & PSNR/SSIM             & PSNR/SSIM             \\   \midrule
Full-ft    & 11.6M & 27.29/0.7800          & 25.58/0.6598          & 23.71/0.6768          & 25.11/0.6096          & 25.69/0.8043          \\
Pretrain   & -     & 25.08/0.6638          & 23.95/0.5847          & 21.51/0.5569          & 24.08/0.5580          & 22.60/0.6612          \\
VPT~\cite{jia2022visual}        & 311K  & 26.39/0.7367          & 24.84/0.6306          & 22.48/0.6101          & 24.75/0.5902          & 23.98/0.7365          \\
Adapter~\cite{houlsby2019parameter}    & 168K  & 27.00/0.7698          & 25.36/0.6518          & 23.30/0.6566          & 25.01/0.6039          & 25.13/0.7848          \\
LoRA~\cite{hu2021lora}       & 155K  & 27.01/0.7694          & 25.36/0.6513          & 23.26/0.6551          & 25.00/0.6038          & 25.09/0.7837          \\
AdaptFor.~\cite{chen2022adaptformer}  & 162K  & 27.03/0.7715          & 25.40/0.6533          & 23.32/0.6581          & 25.02/0.6048          & 25.19/0.7873          \\
SSF~\cite{lian2022scaling}        & 117K  & 26.91/0.7664          & 25.33/0.6502          & 23.21/0.6519          & 24.98/0.6027          & 24.98/0.7801          \\
FacT~\cite{jie2023fact}       & 174K  & 27.01/0.7703          & 25.37/0.6521          & 23.30/0.6569          & 25.00/0.6041          & 25.14/0.7855          \\
Ours       & 170K  & \textbf{27.13/0.7739} & \textbf{25.44/0.6545} & \textbf{23.41/0.6620} & \textbf{25.04/0.6057} & \textbf{25.29/0.7903} \\ \bottomrule
\end{tabular}%
}
\label{table:more-edt}
\end{table}

\begin{wrapfigure}{r}{0.40\linewidth}
    \centering
    \vspace{-3mm}
    \captionof{table}{Results on real-world denoising tasks with SIDD datasets.}
    \label{tab:suppl-real-denoise}
    \setlength{\tabcolsep}{2.5pt}
    \scalebox{0.68}{
    \begin{tabular}{@{}l|cccccc@{}}
    \toprule
    Methods & AdaptFor. & LoRA  & Adapter & FacT  & MoE   & Ours  \\ \midrule
    \#param & 677K      & 995K  & 691K    & 537K  & 667K  & 697K  \\
    PSNR    & 39.03     & 38.97 & 39.00   & 39.02 & 39.05 & 39.10 \\ \bottomrule
    \end{tabular}%
    }
\end{wrapfigure}

\subsection{Results on More Degradations.}

In this section, we further include another challenging degradation type, namely the real image denoising which is unseen during the pre-training phase and is the real-world degradation type, to further demonstrate the generalization of the proposed AdaptIR. We use the training and testing sets in the SIDD for this experiment. The experimental results are shown in~\cref{tab:suppl-real-denoise}. It can be seen that our AdaptIR maintains its superiority when transferring to real-world degradation. For instance, our method outperforms LoRA by 0.13dB PSNR. The above experimental results demonstrate the robustness of our methods.

\section{Complexity Analysis}
\label{sec:param-analysis}
In this section, we theoretically analyze the parameter complexity of the proposed method. We omit the bias term as the corresponding parameter is small. Assume that the hidden dimension of the pre-trained restoration model~\cite{chen2021pre,li2021efficient} is $d$ and the dimension of intrinsic space in AdaptIR is $d'$. For the dimensional up and down operations, the number of parameters is $2dd'$. For the Local Interaction Module, assuming the convolution kernel size is $K$ and the pre-defined rank of $U$, $V$ is $r$, then the number of parameters of LIM with depth-separable design is $d'r+rK^2$. For the Frequency Affine Module, the total parameters of the amplitude and phase projection are $2d'$. For the Channel Gating Module, which contains the channel compression as well as the FFN, the number of parameters is $d'+2d'\frac{d'}{a}$. As for the Adaptive Feature Ensemble, the compression convolution costs $d'$ number of parameters, and $2d'\frac{d'}{b}$ is used in pooling FFN. Summing up the above terms gives $\frac{2(a+b)}{ab}d'^2 + (r+4+2d)d'+rK^2$. In the implementation, we set $r=d'/2$, $a=2$,$b=8$, $K=3$ and $d'=\frac{d}{\gamma}$. Therefore, the total parameter complexity of AdaptIR is $(\frac{2}{\gamma}+\frac{7}{4\gamma^2})d^2+\frac{17}{2\gamma}d \sim \mathcal{O}(\frac{d^2}{\gamma})$.

\begin{figure*}[h]
    \centering
    \includegraphics[width=0.98\linewidth]{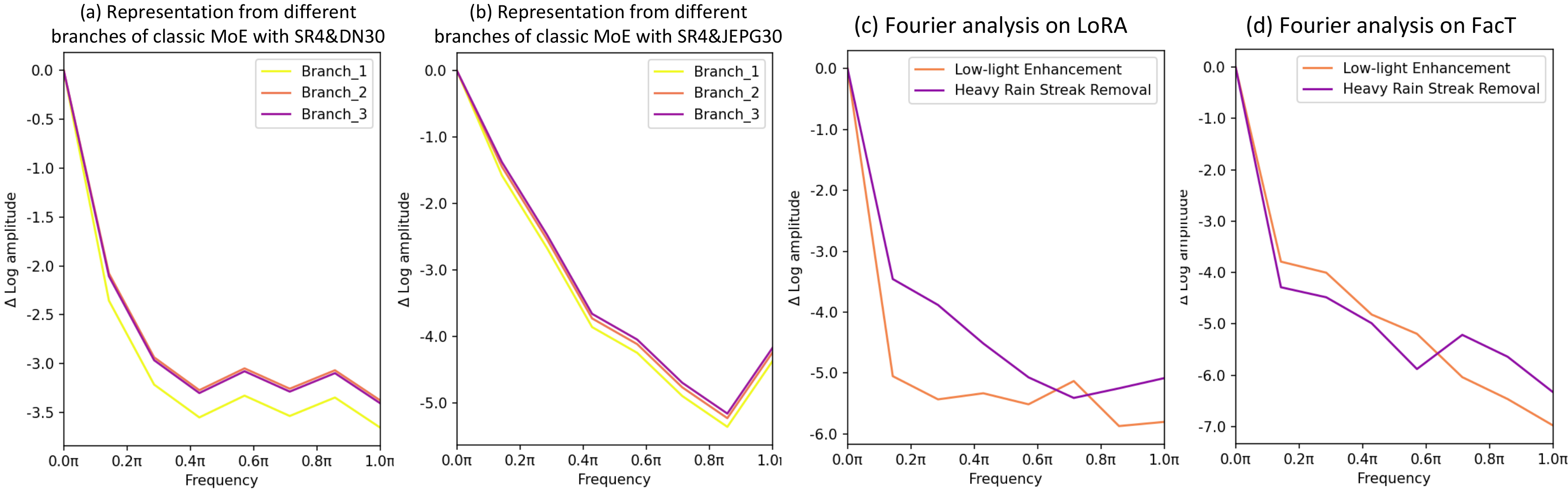}
    \caption{The frequency characteristic curves of features from three branches in the classic MoE with hybrid degradation of (a)SR4\&DN30 and (b)SR4\&JPEG30. (c)\&(d) Fourier analysis on more current PETL methods, LoRA~\cite{hu2021lora} and FacT~\cite{jie2023fact}, which shows significant representation homogeneity across tasks.
    }
    \label{fig:difference-moe-other-observation}
\end{figure*}

\begin{figure*}[h]
    \centering
    \includegraphics[width=0.98\linewidth]{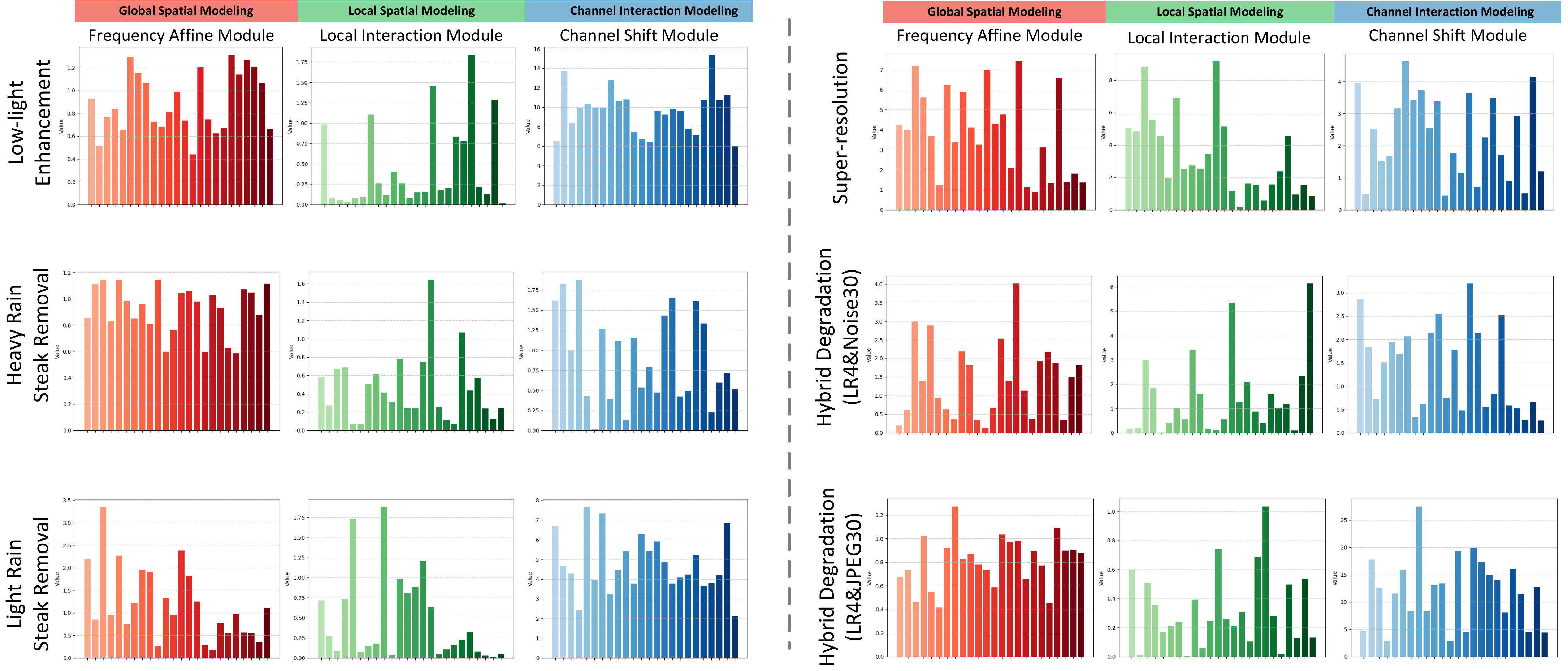}
    \caption{The distribution of feature response densities of the three branches across different tasks.}
    \label{fig:suppl-distribution-of-three-branch}
\end{figure*}

\section{Feature Response Intensity Analysis.}
To make it more clear how the proposed multi-branch structure works, we give the distribution of feature response intensity of three branches across various tasks, including SR, heavy deraining, light deraining, low-light image enhancement, and two hybrid degradations in~\cref{fig:suppl-distribution-of-three-branch}. These figures indicate that our AdaptIR can adjust to different degradation types by enhancing or suppressing the outputs from different branches. Specifically, for the heavy\&light deraining tasks, AdaptIR adaptively learns to enhance the low-frequency global features, i.e., the frequency affine module which is responsible for global spatial modeling has large values. This property ensures the removal of the high-frequency rainstreaks as well as the preservation of the global structure of the image. For SR tasks, AdaptIR adaptively enhances the restoration of local texture details by learning large output values from the local spatial modules. For the hybrid degradation task, AdaptIR shows it can distinguish between different hybrid degradations, i.e., three branches exhibit different patterns under two types of hybrid degradations. In short, each branch of AdaptIR can capture discriminative features under different degradations, indicating that our approach is degradation-aware. This ability guarantees robustness on single degradation and superior performance under hybrid degradation.

\section{Differences from Classic MoE}
\label{sec:diff-moe}
Although both our AdapIR and the classic MoE employ the multi-branch structure, however, our approach differs from the classic MoE in the following aspects. \textbf{Firstly}, the classic MoE uses the multi-branch structure to enhance the model capabilities, whereas our proposed heterogeneous MoE aims to capture heterogeneous representations across different restoration tasks. \textbf{Secondly}, despite using the multi-branch structure, the classical MoE still tends to capture homogeneous representations since each branch is the same, thus resulting in the sub-optimal results in~\cref{table:compare-hybrid}. In contrast, each branch in our AdaptIR is designed orthogonally, thus ensuring the learning of orthogonal representation bases. \textbf{Thirdly}, classical MoE uses simple summation to fuse branches, which is degradation-agnostic, while our AdaptIR uses degradation-specific ensemble to learn the combination of orthogonal representation bases, facilitating heterogeneous representation across tasks.

In~\cref{fig:difference-moe-other-observation}, we also give the frequency characteristics of the output features from different branches of the well-trained classical MoE. It can be seen that different branches still suffer from homogeneity despite the use of a multi-branch structure. In contrast, as shown in~\cref{fig:fourior} in the main paper, our AdaptIR ensures that different branches capture different representations through the proposed orthogonal design, which promotes heterogeneous representations to achieve better performance.

\section{More Evidence of Homogeneous Representation}
\label{sec:more-evidence}
In~\cref{fig:motivation}, we give the frequency characteristics of Adapter~\cite{lester2021power}, and find its homogeneous representation when facing different degradations. To demonstrate the prevalence of homogeneous representations in current PETL methods, we provide the frequency characteristics curves of more PETL methods in~\cref{fig:difference-moe-other-observation}. It can be seen that the current state-of-the-art PETL methods LoRA~\cite{hu2021lora} and FacT~\cite{jie2023fact} also exhibit homogeneity as Adapter, \textit{i.e.}, the learned feature representations are similar even if they are for different degradations. In contrast, AdapIR utilizes the orthogonal multi-branch design to learn diverse representations, facilitating heterogeneous representations on different restoration tasks.

\begin{table}[!tb]
\centering
\caption{Dataset description for various image restoration tasks.}
\label{tab:dataset-description}
\setlength{\tabcolsep}{4pt}
\scalebox{0.9}{
\begin{tabular}{@{}lccc@{}}
\toprule
Tasks                    & Type  & Dataset                              & Num\_samples     \\ \midrule
\multirow{2}{*}{Super-resolution}      & train & Div2K+Flicker2K                      & 800+2650         \\
                         & test  & Set5+Set14+BSDS100+Urban100+Manga109 & 5+14+100+100+109 \\ \midrule
\multirow{2}{*}{Denoise} & train & BSD400+WED                           & 400+4744         \\
                         & test  & BSD68+Urban100                       & 68+100           \\ \midrule
\multirow{2}{*}{DerainL} & train & RainTrainL                           & 200              \\
                         & test  & Rain100L                             & 100              \\ \midrule
\multirow{2}{*}{DerainH} & train & RainTrainH                           & 1800             \\
                         & test  & Rain100H                             & 100              \\ \midrule
\multirow{2}{*}{\begin{tabular}[c]{@{}l@{}}Second-order Restor. \\ (SR4\&Dnoise30)\end{tabular}} & train & Div2K+Flicker2K   & 800+2650 \\
                         & test  & Set5+Set14+BSDS100+Urban100+Manga109 & 5+14+100+100+109 \\ \midrule
\multirow{2}{*}{\begin{tabular}[c]{@{}l@{}}Low-light \\ Enhancement\end{tabular}}      & train & LOLv1-train-split & 485      \\
                         & test  & LoLv1-test-split                     & 15               \\ \bottomrule
\end{tabular}%
}
\end{table}

\section{Dataset Description}
In this work, we evaluate different PETL methods on diverse image restoration tasks, which cover many training and testing datasets. To make the experimental setup more clear, we give a detailed description of datasets in \cref{tab:dataset-description}.

\section{Limitations and Future Work}
\label{sec:limitation}
While AdaptIR appears as a competitive PETL alternative across various image restoration benchmarks, it can be further improved with task-specific module designs. For example, in the proposed AdaptIR, different tasks share the same structure, however, different restoration tasks have diverse model preferences. An intuitive solution might be to introduce degradation-aware dynamic networks. Moreover, although this work has covered multiple degradation types, some other degradations can also be explored in the future, \textit{e.g.} blur and haze, to further demonstrate the generalization ability.

\section{Broader Impact}
\label{sec:broader-impact}
Our AdaptIR holds significant promise for improving the quality and generalization of image restoration across various domains, such as medical imaging, historical document preservation, and digital media restoration. By enabling more accurate and reliable image restoration with reduced computational resources, AdaptIR can facilitate advancements in these fields, leading to better diagnostic tools, preservation of cultural heritage, and enhanced digital media quality. However, the enhanced capabilities of AdaptIR also present potential negative societal impacts, such as the risk of misuse in generating realistic fake images or deepfakes, which could be used for disinformation, creating fake profiles, or unauthorized surveillance, leading to privacy violations, security concerns, and ethical issues. To mitigate these risks, it is crucial to implement measures like gated releases of models, mechanisms for monitoring misuse and ensuring transparency in deployment and training processes, alongside continuous evaluation of the technology's impact.

\end{document}